\documentclass[12pt]{spieman}  
\usepackage{graphicx}
\usepackage{setspace}
\usepackage{tocloft}
\usepackage{booktabs}
\usepackage{cite}
\usepackage[cmex10]{amsmath}
\usepackage{algorithm}
\usepackage{algpseudocode}
\usepackage{float}
\usepackage{mdwmath}
\usepackage{mdwtab}
\usepackage{array}
\usepackage{booktabs}
\usepackage{multirow}
\usepackage{flushend}
\usepackage{setspace}
\usepackage{tabularx}
\usepackage{rotating}
\usepackage{longtable}

\usepackage{color}

\newcommand{\ParSection}[1]{}

\newcommand{\ParSubSection}[1]{}

\title{A Comprehensive Survey of Deep Learning in Remote Sensing: Theories, Tools and Challenges for the Community}

\author[a,*]{John E. Ball}
\author[a]{Derek T. Anderson}
\author[b]{Chee Seng Chan}

\affil[a]{Mississippi State University, Department of Electrical and Computer Engineering, 406 Hardy Rd., Mississippi State, MS, USA, 39762}
\affil[b]{University of Malaya, Faculty of Computer Science and Information Technology, 50603 Lembah Pantai, Kuala Lumpur, Malaysia}

\cftpagenumbersoff{figure}
\cftpagenumbersoff{table}

\begin{document}


\maketitle

\begin{abstract}
In recent years, deep learning (DL), a re-branding of neural networks (NNs), has risen to the top in numerous areas, namely computer vision (CV), speech recognition, natural language processing, etc. Whereas remote sensing (RS) possesses a number of unique challenges, primarily related to sensors and applications, inevitably RS draws from many of the same theories as CV; e.g., statistics, fusion, and machine learning, to name a few. This means that the RS community should be aware of, if not at the leading edge of, of advancements like DL. Herein, we provide the most comprehensive survey of state-of-the-art RS DL research. We also review recent new developments in the DL field that can be used in DL for RS. Namely, we focus on theories, tools and challenges for the RS community. Specifically, we focus on unsolved challenges and opportunities as it relates to (i) inadequate data sets, (ii) human-understandable solutions for modelling physical phenomena, (iii) Big Data, (iv) non-traditional heterogeneous data sources, (v) DL architectures and learning algorithms for spectral, spatial and temporal data, (vi) transfer learning, (vii) an improved theoretical understanding of DL systems, (viii) high barriers to entry, and (ix) training and optimizing the DL.

\end{abstract}


{\noindent \footnotesize\textbf{*}John E. Ball, \linkable{jeball@ece.msstate.edu} }

%
%
\begin{spacing}{1}    

%
%
\section{Introduction}
\label{sec:Introduction}

\ParSection{Motivation (DEREK-DONE)} In recent years, deep learning (DL) has led to leaps, versus incremental gain, in fields like computer vision (CV), speech recognition, and natural language processing, to name a few. The irony is that DL, a surrogate for neural networks (NNs), is an age old branch of artificial intelligence that has been resurrected due to factors like algorithmic advancements, high performance computing, and Big Data. The idea of DL is simple; the machine is learning the features and decision making (classification), versus a human manually designing the system. The reason this article exists is remote sensing (RS). The reality is, RS draws from core theories such as physics, statistics, fusion, and machine learning, to name a few. This means that the RS community should be aware of, if not at the leading edge of, advancements like DL. The aim of this article is to provide resources with respect to theory, tools and challenges for the RS community. Specifically, we focus on unsolved challenges and opportunities as it relates to (i) inadequate data sets, (ii) human-understandable solutions for modelling physical phenomena, (iii) Big Data, (iv) non-traditional heterogeneous data sources, (v) DL architectures and learning algorithms for spectral, spatial and temporal data, (vi) transfer learning, (vii) an improved theoretical understanding of DL systems, (viii) high barriers to entry, and (ix) training and optimizing the DL.

\ParSection{Define Remote Sensing (JOHN-DONE)} Herein, RS is a technological challenge where objects or scenes are analyzed by remote means. This includes the traditional remote sensing areas, such as satellite-based and aerial imaging. This definition also includes non-traditional areas, such as unmanned aerial vehicles (UAVs), crowdsourcing (phone imagery, tweets, etc.), advanced driver assistance systems (ADAS), etc. These types of remote sensing offer different types of data and have different processing needs, and thus also come with new challenges to algorithms that analyze the data.

\ParSection{Paper Contributions (JOHN-DONE)} The contributions of this paper are as follows:\\

%
%
\begin{enumerate}

\item \textit{Thorough list of challenges and open problems in DL RS}. We focus on unsolved challenges and opportunities as it relates to (i) inadequate data sets, (ii) human-understandable solutions for modelling physical phenomena, (iii) Big Data, (iv) non-traditional heterogeneous data sources, (v) DL architectures and learning algorithms for spectral, spatial and temporal data, (vi) transfer learning, (vii) an improved theoretical understanding of DL systems, (viii) high barriers to entry, and (ix) training and optimizing the DL. These observations are based on surveying RS DL and feature learning literature, as well as numerous RS survey papers. This topic is the majority of our paper and is discussed in section \ref{sec:ChallengesOpportunitiesDLRS}.

\item \textit{Thorough literature survey}. Herein, we review 207 RS application papers, and 57 survey papers in remote sensing and DL. In addition, many relevant DL papers are cited. Our work extends the previous DL survey papers \cite{Cheng, Deng2014, Zhang2016} to be more comprehensive. We also cluster DL approaches into different application areas and provide detailed discussions of some relevant example papers in these areas in section \ref{sec:DL_Approaches_RS}.

\item \textit{Detailed discussions of modifying DL architectures to tackle RS problems.} We highlight approaches in DL in RS, including new architectures, tools and DL components, that current RS researchers have implemented in DL. This is discussed in section \ref{subsec:ChallengesOpportunities_v}.

\item \textit{Overview of DL}. For RS researchers not familiar with DL, section \ref{sec:DeepShallowLearning} provides a high-level overview of DL and lists many good references for interested readers to pursue.

\item \textit{Deep learning tool list}. Tools are a major enabler of DL, and we review the more popular DL tools. We also list pros and cons of several of the most popular toolsets and provide a table summarizing the tools, with references and links (refer to Table \ref{table:DLTools}). For more details, see section \ref{subsubsec:Tools}.

\item \textit{Online summaries of RS datasets and DL RS papers reviewed}. First, an extensive online table with details about each DL RS paper we reviewed: sensor modalities, a compilation of the datasets used, a summary of the main contribution, and references. Second, a dataset summary for all the DL RS papers analyzed in this paper is provided online. It contains the dataset name, a description, a URL (if one is available) and a list of references. Since the literature review for this paper was so extensive, these tables are too large to put in the main article, but are provided online for the readers' benefit. These tables are located at \url{http://www.cs-chan.com/source/FADL/Online_Dataset_Summary_Table.pdf} and \url{http://www.cs-chan.com/source/FADL/Online_Paper_Summary_Table.pdf}.

\end{enumerate}

\ParSection{Short table of acronyms (JOHN-ALMOST DONE)}

As an aid to the reader, Table \ref{table:Acronyms} lists acronyms used in this paper.

\begin{center}
\begin{longtable}{|p{1.5cm}|p{5.5cm}|p{1.5cm}|p{5.5cm}|}
\caption{Acronym list.} 
\label{table:Acronyms}

\\
\hline
\multicolumn{1}{|c|}{\textbf{Acronym}} & 
\multicolumn{1}{c|}{\textbf{Meaning}} & 
\multicolumn{1}{c|}{\textbf{Acronym}} & 
\multicolumn{1}{c|}{\textbf{Meaning}} \\ 
\hline 
\endfirsthead

\multicolumn{4}{c}%
{{\bfseries \tablename\ \thetable{} -- continued from previous page}} \\
\hline 
\multicolumn{1}{|c|}{\textbf{Acronym}} & 
\multicolumn{1}{c|}{\textbf{Meaning}} & 
\multicolumn{1}{c|}{\textbf{Acronym}} & 
\multicolumn{1}{c|}{\textbf{Meaning}} \\ 
\hline 
\endhead

\hline 
\multicolumn{4}{|c|}{{Continued on next page}} \\
\hline
\endfoot

\hline
\hline
\endlastfoot

ADAS      & Advanced Driver Assistance System &                        AE        & AutoEncoder   \\  \hline  
ANN       & Artificial Neural Network &                                ATR       & Automated Target Recognition \\  \hline
AVHRR     & Advanced Very High Resolution Radiometer &                 AVIRIS    & Airborne Visible / Infrared Imaging Spectrometer \\  \hline
BP        & Backpropagation &                                          CAD       & Computer Aided Design \\  \hline
CFAR      & Constant False Alarm Rate &                                CG        & Conjugate Gradient \\  \hline
ChI       & Choquet Integral &                                         CV        & Computer Vision \\  \hline
CNN       & Convolutional Neural Network &                             DAE       & Denoising AE      \\  \hline
DAG       & Directed Acyclic Graph &                                   DBM       & Deep Boltzmann Machine \\  \hline
DBN       & Deep Belief Network &                                      DeconvNet      & DeConvolutional Neural Network \\  \hline
DEM       & Digital Elevation Model &                                  DIDO      & Decision In Decision Out \\  \hline
DL        & Deep Learning &                                            DNN       & Deep Neural Network \\  \hline
DSN       & Deep Stacking Network &                                    DWT       & Discrete Wavelet Transform \\  \hline
FC        & Fully Connected &                                          FCN       & Fully Convolutional Network \\  \hline
FC-CNN    & Fully Convolutional CNN &                                  FC-LSTM   & Fully Connected LSTM \\  \hline
FIFO      & Feature In Feature Out &                                   FL        & Feature Learning \\  \hline  
GBRCN     & Gradient-Boosting Random Convolutional Network &           GIS       & Geographic Information System \\  \hline
GPU       & Graphical Processing Unit &                                HOG       & Histogram of Ordered Gradients \\  \hline
HR        & High Resolution &                                          HSI       & HyperSpectral Imagery \\  \hline
ILSVRC    & ImageNet Large Scale Visual Recognition Challenge &        L-BGFS    & Limited Memory BGFS \\  \hline
LBP       & Local Binary Patterns &                                    LiDAR     & Light Detection and Ranging \\  \hline
LR        & Low Resolution &                                           LSTM      & Long Short-Term Memory \\  \hline
LWIR      & Long-Wave InfraRed &                                       MKL       & Multi-Kernel Learning \\  \hline
MLP       & Multi-Layer Perceptron &                                   MSDAE     & Modified Sparse Denoising Autoencoder \\  \hline
MSI       & MultiSpectral Imagery   &                                  MWIR      & Mid-wave InfraRed \\  \hline
NASA      & National Aeronautics and Space Administration &            NN        & Neural Network \\  \hline
NOAA      & National Oceanic and Atmospheric Administration &          PCA       & Principal Component Analysis \\  \hline
PGM       & Probabilistic Graphical Model &                            PReLU     & Parametric Rectified Linear Unit \\  \hline
RANSAC    & RANdom SAmple Concesus &                                   RBM       & Restricted Boltzmann Machine   \\  \hline
ReLU      & Rectified Linear Unit &                                    RGB       & Red, Green and Blue image \\  \hline
RGBD      & RGB + Depth image &                                        RF        & Receptive Field \\  \hline
RICNN     & Rotation Invariant CNN &                                   RNN       & Recurrent NN \\  \hline
RS        & Remote Sensing &                                           R-VCANet  & Rolling guidance filter Vertex Component Analysis NETwork \\  \hline
S-MSDAE   & Stacked MSDAE &                                            SAE       & Stacked AE \\  \hline
SAR       & Synthetic Aperture Radar &                                 SDAE      & Stacked DAE \\  \hline
SIDO      & Signal In Decision Out &                                   SIFT      & Scale Invariant Feature Transform \\  \hline
SISO      & Signal In Signal Out &                                     SGD       & Stochastic Gradient Descent  \\  \hline
SPI       & Standardized Precipitation Index &                         SSAE      & Stacked Sparse Autoencoder \\  \hline
SVM       & Support Vector Machine &                                   UAV       & Unmanned Aerial Vehicle \\  \hline
VGG       & Visual Geometry Group &                                    VHR       & Very High Resolution \\  \hline

\end{longtable}
\end{center}

\ParSection{Overview of paper (JOHN-DONE)} 
This paper is organized as follows. Section \ref{sec:DeepShallowLearning} discusses related work in CV. This section contrasts deep and ``shallow'' learning, and discusses DL architectures. The main reasons for success of DL are also discussed in this section. Section \ref{sec:DL_Approaches_RS} provides an overview of DL in RS, highlighting DL approaches in many disparate areas of RS. Section \ref{sec:ChallengesOpportunitiesDLRS} discusses the unique challenges and open issues in applying DL to RS. Conclusions and recommendations are listed in section \ref{sec:Conclusions}.

%
%
\section{Related work in CV}
\label{sec:DeepShallowLearning}

\ParSection{Introduction (JOHN-DONE)}
CV is a field of study trying to achieve visual understanding through computer analysis of imagery. In the past, typical approaches utilized a processing chain which usually started with image denoising or enhancement, followed by feature extraction (with human coded features), a feature optimization stage, and then processing on the extracted features. These architectures were mostly ``shallow'', in the sense that they usually had only one to two processing layers between the features and the output. Shallow learners (Support Vector Machines (SVMs), Gaussian Mixture Models, Hidden Markov Models, Conditional Random Fields, etc.) have been the backbone of traditional research efforts for many years \cite{Deng2014} . In contrast, DL usually has many layers (the exact demarcation between ``shallow'' and ``deep'' learning is not a set number), which allows a rich variety of highly complex, nonlinear and hierarchical features to be learned from the data. The following sections contrast deep and shallow learning, discuss DL approaches and DL enablers, and finally discuss DL success in domains other than RS.

\subsection{Deep vs. shallow learning}

\textit {Shallow learning} is a term used to describe learning networks that usually have at most one to two layers. Examples of shallow learners include the popular SVM, Gaussian mixture models, hidden Markov models, conditional random fields, logistic regression models, and the extreme learning machine \cite{Deng2014} . Shallow learning models usually have one or two layers that compute a linear or non-linear function of the input data (often hand-designed features). DL, on the other hand, usually means a deeper network, with many layers of (usually) non-linear transformations. Although there is no universally accepted definition of how many layers constitute a ``deep'' learner, typical networks are typically at least four or five layers deep. Three main features of DL systems are that DL systems (1) can learn features directly from the data itself, versus human-designed features, (2) can learn hierarchical features which increase in complexity through the deep network, and (3) can be more generalizable and more efficiently encode the model compared to a shallower NN approach; that is, a shallow system will require exponentially more neurons (and thus more free parameters) and more training data \cite{wang2015survey, Wan2014Deep} . An interesting study on deep and shallow nets is given by Ba and Caruana \cite{ba2014deep} , where they perform model compression, by training a Deep NN (DNN). The unlabeled data is then evaluated by the DNN and the scores produced by that model are used to train a compressed (shallower) model. If the compressed model learns to mimic the large model perfectly it makes exactly the same predictions and mistakes as the complex model. The key is the compressed model has to have enough complexity to regenerate the more complex model output.

DL systems are often designed to loosely mimic human or animal processing, in which there are many layers of interconnected components, e.g. human vision. So there is a natural motivation to use deep architectures in CV-related problems. For the interested reader, we provide some useful survey paper references. Arel et al. provide a survey paper on DL \cite{Arel2010} . Deng et al. \cite{Deng2014} provide two important reasons for DL success: (1) Graphical Processing Unit (GPU) units and (2) recent advances in DL research. They discuss generative, discriminative, and hybrid deep architectures and show there is vast room to improve the current optimization techniques in DL. Liu et al. \cite{liu2016survey} give an overview of the autoencoder, the CNN, and DL applications. Wang et al. provide a history of DL \cite{wang2015survey} . Yu et al. \cite{Yu2011} provide a review of DL in signal and image processing. Comparisons are made to shallow learning, and DL advantages are given. Two good overviews of DL are the survey paper of Schmidhuber et al. \cite{Schmidhuber2015} and the book by Goodfellow et al. \cite{goodfellow2016deep}. Zhang et al. \cite{Zhang2016} give a general framework for DL in remote sensing, which covers four RS perspectives: (1) image processing, (2) pixel-based classification, (3) target recognition, and (4) scene understanding. In addition, they review many DL applications in remote sensing. Cheng et al. discuss both shallow and DL methods for feature extraction \cite{Cheng}. Some good DL papers are the introductory DL papers of Arnold et al. \cite{Arnold2012Introduction} and Wang et al. \cite{wang2015survey} , the DL book by Goodfellow et al. \cite{goodfellow2016deep} , and the DL survey papers \cite{Schmidhuber2015, bengio2012unsupervised, Chen2014, Deng2013Deep, Deng2014, liu2016survey, Najafabadi2015Deep, wang2015survey, Wan2014Deep} .

\subsection{Traditional Feature Learning methods}
Traditional methods of feature extraction involve hand-coded features to extract information based on spatial, spectral, textural, morphological content, etc. These traditional methods are discussed in detail in the following references, and we will not give extensive algorithmic details herein. All of these hand-derived features are designed for a specific task, e.g. characterizing image texture. In contrast, DL systems derive complicated, (usually) non-linear and hierarchical features from the data itself. 

Cheng et al. \cite{Cheng} discuss traditional handcrafted features such as the Histogram of Ordered Gradients (HOG), the Scale-Invariant Feature Transform (SIFT) and SIFT variants, color histograms, etc. They also discuss unsupervised FL methods, such as principal components analysis, $k$-means clustering, sparse coding, etc. Other good survey papers discuss hyperspectral image (HSI) data analysis \cite{Bioucas-dias2013} , kernel-based methods \cite{Camps-Valls2005Kernel} , statistical learning methods in HSI \cite{camps2013advances} , spectral distance functions \cite{deborah2015comprehensive} , pedestrian detection \cite{dollar2012pedestrian} , multi-classifier systems \cite{du2012multiple} , spectral-spatial classification \cite{fauvel2013advances} , change detection \cite{hussain2013change, jianya2008review} , machine learning in RS \cite{lary2016machine} , manifold learning \cite{lunga2014manifold} , transfer learning \cite{pan2010survey} , endmember extraction \cite{plaza2004quantitative} , and spectral unmixing \cite{keshava2002spectral, keshava2003survey, parente2010survey, plaza2011recent, shi2014incorporating} .

\subsection{DL Approaches}
To date, the auto-encoder (AE), the CNN, Deep Belief Networks (DBNs), and the Recurrent NN (RNN), have been the four mainstream DL architectures. The deconvolutional NN (DeconvNet) is a relative newcomer to the DL community. The following sections discuss each of these architectures at a high level. Many good references are provided for the interested reader.

\subsubsection{Autoencoder (AE)}
An AE is a network designed to learn useful features from unsupervised data. One of the first applications of AEs was dimensionality reduction, which is required in many RS applications. By reducing the size of the adjacent layers, the AE is forced to learn a compact representation of the data. The AE maps the input through an encoder function $f$ to generate an internal (latent) representation, or code, $h$, that is, $h = f(\textbf{x})$. The autoencoder also has a decoder function, $g$ that maps $h$ to the output $\hat{\textbf{x}}$. In general, the AE is constrained, either through its architecture, or through a sparsity constraint (or both), to learn a useful mapping (but not the trivial identity mapping). A loss function $L$ measures how close the AE can reconstruct the output: $L$ is a function of $\textbf{x}$ and $\hat{\textbf{x}} = g(f(\textbf{x}))$. A regularization function $\Omega(h)$ can also be added to the loss function to force a more sparse solution. The regularization function can involve penalty terms for model complexity, model prior information, penalizing based on derivatives, or penalties based on some other criteria such as supervised classification results, etc. (reference \S 14.2 of \cite{goodfellow2016deep} ). 

\begin{figure*}
\begin{center}
 \includegraphics[height=.4\textheight]{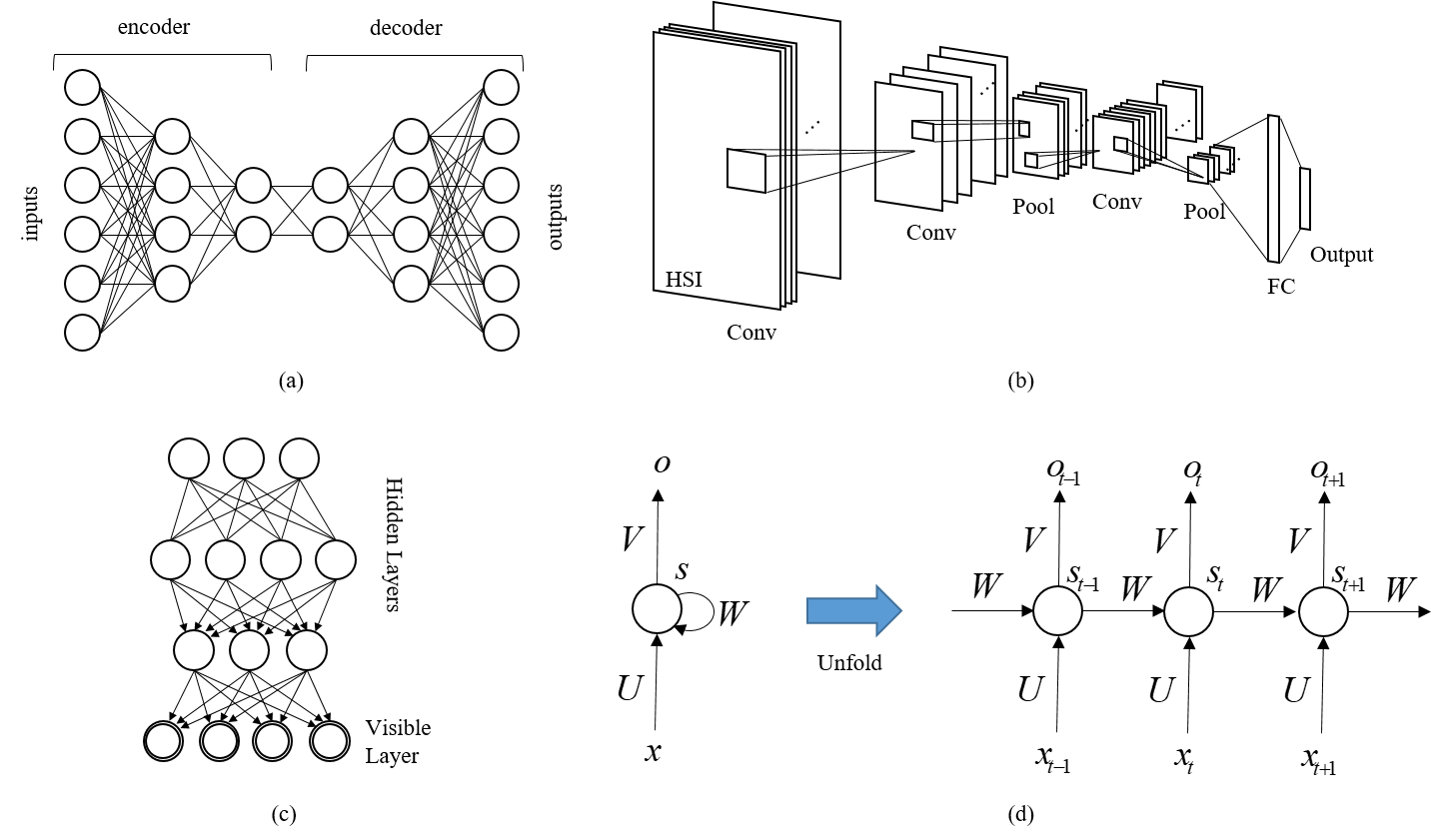}
 \caption{Block diagrams of DL architectures. (a) AE. (b) CNN. (c) DBN. (d) RNN.}
 \label{fig:DLBlockDiagrams}
\end{center}
\end{figure*}

A Denoising AE (DAE) is an AE designed to remove noise from a signal or an image. Chen et al. developed an efficient DAE, which marginalizes the noise and has a computationally efficient closed form solution \cite{chen2012marginalized}. To provide robustness, the system is trained using additive Gaussian noise or binary masking noise (force some percentage of inputs to zero). Many RS applications utilize an AE for denoising. Figure \ref{fig:DLBlockDiagrams}(a) shows an example of a AE. The diabolo shape results in dimensionality reduction.

\subsubsection{Convolutional Neural Network (CNN)}
A CNN is a network that is loosely inspired by the human visual cortex. A typical CNN is composed of multiple dual-layers of convolutional masks followed by pooling, and these layers are then usually followed by either fully-connected or partially-connected layers, which perform classification or class probability estimation. Some CNNs also utilize data normalization layers. The convolution masks have coefficients that are learned by the CNN. A CNN that analyzes grayscale imagery will employ 2D convolution masks, while a CNN using Red-Green-Blue (RGB) imagery will use 3D masks. Through training, these masks learn to extract features directly from the data, in stark contrast to traditional machine learning approaches, which utilize "hand-crafted" features. The pooling layers are non-linear operators (usually maximum operators), which allows the CNN to learn non-linear features, which greatly increases its learning capabilities. Figure \ref{fig:DLBlockDiagrams}(b) shows an example CNN, where the input is a HSI, and there are two convolution and pooling layers, followed by two fully connected (FC) layers.

The number of convolution masks, the size of the masks, and the pooling functions are all parameters of the CNN. The masks at the first layers of the CNN typically learn basic features, and as one traverses the depths of the network, the features become more complex and are built-up hierarchically. Normalization layers provide regularization and can aid in training. The fully-connected layers (or partially-connected layers) are usually near the end of the CNN network, and allow complex non-linear functions to be learned from the hierarchical outputs of the previous layers. These final layers typically output class labels or estimates of the probabilities of the class label.

CNNs have dominated in many perceptual tasks. Following Ujjwalkarn \cite{Ujjwalkarn2016Intuitive} , the image recognition community has shown keen interest in CNNs. Starting in the 1990s, LeNet was developed by LeCun et al. \cite{lecun1998gradient} , and was designed for reading zip codes. It generated great interest in the image processing community. In 2012, Krizhevsky et al. \cite{krizhevsky2012imagenet} introduced AlexNet, a deep CNN. It won the ImageNet Large Scale Visual Recognition Challenge (ILSVRC) in 2012 by a significant margin. In 2013, Zeiler and Fergus \cite{zeiler2014visualizing} created ZFNet, which was AlexNet with tweaked parameters and won ILSVRC. Szegedy et al. \cite{szegedy2015going} won ILSVRC with GoogLeNet in 2014, which used a much smaller number of parameters (4 million) than AlexNet (60 million). In 2015, ResNets were developed by He et al \cite{he2016deep} , which allowed CNNs to have very deep networks. In 2016, Huang et al. \cite{huang2016densely} published DenseNet, where each layer is directly connected to every other layer in a feedforward fashion. This architecture also eliminates the vanishing-gradient problem, allowing very deep networks. The examples above are only a few examples of CNNs.

\subsubsection{Deep Belief Network (DBN)}
A DBN is a type (generative) of Probabilistic Graphical Model (PGM), the marriage of probability and graph theory. Specifically, a DBN is a ``deep'' (large) Directed Acyclic Graph (DAG). A number of well-known algorithms exist for exact and approximate inference (infer the states of unobserved (hidden) variables) and learning (learn the interactions between variables) in PGMs. A DBN can also be thought of as a type of deep NN. In \cite{HinSal06} , Hinton showed that a DBN can be viewed and trained (in a greedy manner) as a stack of simple unsupervised networks, namely Restricted Boltzmann Machines (RBMs), or generative AEs. To date, CNNs have demonstrated better performance on various benchmark CV data sets. However, in theory DBNs are arguably superior. CNNs possess generally a lot more ``constraints''. The DBN versus CNN topic is likely subject to change as better algorithms are proposed for DBN learning. Figure \ref{fig:DLBlockDiagrams}(c) depicts a DBN, which is made up of RBM layers and a visible layer.

\subsubsection{Recurrent Neural Network (RNN)}
The RNN is a network where connections form directed cycles. The RNN is primarily used for analyzing non-stationary processes such as speech and time-series analysis. The RNN has memory, so the RNN has persistence, which the AE and CNN don't possess. A RNN can be unrolled and analyzed as a series of interconnected networks that process time-series data. A major breakthrough for RNNs was the seminal work of Hochreiter and Schmidhuber \cite{hochreiter1997long} , the long short-term memory (LSTM) unit, which allows information to be written to a cell, output from the cell, and stored in the cell. The LSTM allows information to flow and helps counteract the vanishing/exploding gradient problems in very deep networks. Figure \ref{fig:DLBlockDiagrams}(d) shows a RNN and its unfolded version.

\subsubsection{Deconvolutional Neural Network (DeconvNet)}
CNNs are often used for classification only. However, a wealth of questions exist beyond classification, e.g., what are our filters really learning, how transferable are these filters, what filters are the most active in a given image, where is a filter the most active at in a given image (or images), or more holistically, where in the image is our object(s) of interest (soft or hard segmentation). To this end, researchers have recently explored deconvolutional NN (DeconvNet) \cite{zeiler2014visualizing,6126474,Zeiler10deconvolutionalnetworks,DBLP:journals/corr/NohHH15} . Whereas CNNs use pooling, which helps us filter noisy activations and address affine transformations, a DeconvNet uses unpooling--the ``inverse'' of pooling. Unpooling makes use of ``switch variables'', which help us place activation in layer $l$ back to its original pooled location in layer $l-1$. Unpooling results in an enlarged, be it sparse, activation map that is fed to deconvolution filters (that are either learned or derived from the CNN filters). In \cite{DBLP:journals/corr/NohHH15} , the Visual Geometry Group (VGG) developed the VGG 16-layer CNN, thus no deconvolution, with its last classification layer removed was used relative to computer vision on non-remotely sensed data. The resultant DeconvNet is twice as large as the VGG CNN. The first part of the network is the VGG CNN and the second part is an architecturally reversed copy of the VGG CNN with pooling replaced by unpooling. The entire network was trained and used for semantic image segmentation. In a different work, Zeiler et al. showed that a DeconvNet can be used to visualize a single CNN filter at any layer or a combination of CNN filters can be visualized for one or more images \cite{6126474,zeiler2014visualizing} . The point is, relevant DeconvNet research exists in the CV literature. 

Two high-level comments are worth noting. First, DeconvNets have been used by some to help rationalize their architecture and operation selections in the context of a visual odyssey of its impact on the filters relative to one another, e.g., existence of a single dominant feature versus a diverse set of features. In many cases its not a rationalization of the final network performance per se, but instead a DeconvNet is a helpful tool that aids them in exploring the vast sea of choices in designing the network. Second, whereas DeconvNet can be used in many cases for segmentation, they do not always produce the segmentation that we might desire. Meaning, if the CNN learned parts, not the full object, then activation of those parts, or a subset thereof, may not equate to the whole and those parts might also be spatially separated in the image. The later makes it challenging to construct a high quality full object segmentation, or segmentation's if there are more than one instance of that object in an image. DeconvNets are basically very recent and have not (yet) been widely adopted by the RS community.

\subsection{DL Meets the Real World}
It is important to understand the different ``factors'' related to the rise and success of DL. This section discusses these factors: GPUs, DL NN expressivness, big data, and tools.  

\subsubsection{GPUs}
GPUs are hardware devices that are optimized for fast parallel processing. GPUs enable DL by offloading computations from the computer's main processor (which is basically optimized for serial tasks) and efficiently performing the matrix-based computations at the heart of many DL algorithms.  The DL community can leverage the personal computer gaming industry, which demands relatively inexpensive and powerful GPUs. A major driver of the research interest in CNNs is the Imagenet contest, which has over one million training images and 1,000 classes  \cite{russakovsky2015imagenet} . DNNs are inherently parallel, utilize matrix operations, and use a large number of floating point operations per second. GPUs are a match because they have the same characteristics \cite{Brown2015Deep} . GPU speedups have been measured at 8.5 to 9 \cite{Brown2015Deep} and even higher depending on the GPU and the code being optimized. The CNN convolution, pooling and activation calculation operations are readily portable to GPUs. 

\subsubsection{DL NN Expressiveness}
Cybenko \cite{Cybenko1989} proved that MLPs are universal function approximators. Specifically, Cybenko showed that a feed-forward network with a single hidden layer containing a finite number of neurons can approximate continuous functions on compact subsets of $\Re^n$, with respect to relatively minimalistic assumptions regarding the activation function. However, Cybenok's proof is an existence theorem, meaning it tells us a solution exists, but it does not tell us how to design or learn such a network. The point is, NNs have an intriguing mathematical foundation that make them attractive with respect to machine learning. Furthermore, in a theoretical work, Telgarsky \cite{telgarsky2016benefits} has shown that for NN with Rectified Linear Units (ReLU) (1) functions with few oscillations poorly approximate functions with many oscillations, and (2) functions computed by NN with few (many) layers have few (many) oscillations. Basically, a deep network allows decision functions with high oscillations. This gives evidence to show why DL performs well in classification tasks, and that shallower networks have limitations with highly oscillatory functions. Sharir et al. \cite{sharir2017expressive} showed that having overlapping local receptive fields and more broadly denser connectivity gives an exponential increase in the expressive capacity of the NN. Liang et al. \cite{liang2016whydeep} showed that shallow networks require exponentially more neurons than a deep network to achieve the level of accuracy for function approximation.   

\subsubsection{Big Data}\label{subsec:bigdatareview}
Every day, approximately $350$ million images are uploaded to Facebook \cite{Brown2015Deep} , Wal-Mart collects approximately $2.5$ petabytes of data per day \cite{Brown2015Deep} , and National Aeronautics and Space Administration (NASA) alone is actively streaming $1.73$ gigabytes of spacecraft borne observation data for active missions alone \cite{Ma201547}. IBM reports that $2.5$ quintillion bytes of data are now generated every data, which means that ``$90\%$ of the data in the world today has been created in the last two years alone'' \cite{7565634} . The point is, an unprecedented amount of (varying quality) data exists due to technologies like remote sensing, smart phones, inexpensive data storage, etc. In times past, researchers used tens to hundreds, maybe thousands of data training samples, but nothing on the order of magnitude as today. In areas like CV, high data volume and variety have been at the heart of advancements in performance. Meaning, reported results are a reflection of advances in both data and machine learning.

To date, a number of approaches have been explored relative to large scale deep networks (e.g., hundreds of layers) and Big Data (e.g., high volume of data). For example, in \cite{Raina:2009:LDU:1553374.1553486} Raina et al. put forth CPU and GPU ideas to accelerate DBNs and sparse coding. They reported a 5 to 15-fold speed up for networks with 100 million plus parameters versus previous works that used only a few million parameters at best. On the other hand, CNNs typically use back propagation and they can be implemented either by pulling of pushing \cite{Ciresan:2011:FHP:2283516.2283603} . Furthermore, ideas like circular buffers \cite{Scherer:2010:EPO:1886436.1886447} and multi GPU based CNN architectures, e.g., Krizhevsky \cite{krizhevsky2012imagenet} , have been put forth. Outside of hardware speedups, operators like ReLUs have been shown to run sever times faster than other common nonlinear functions. In \cite{6288333} , Deng et al. put forth a Deep Stacking Network (DSN) that consists of specialized NNs (called modules), each of which have a single hidden layer. Hutchinson et al. put forth Tensor-DSN is an efficient and parallel extension of DSNs for CPU clusters \cite{Hutchinson:2013:TDS:2498740.2498878} . Furthermore, DistBelief is a library for distributed training and learning of deep networks with large models (billions of parameters) and massive sized data sets \cite{Dean:2012:LSD:2999134.2999271} . DistBelief makes use of machine clusters to manage the data and parallelism via methods like multi-threading, message passing, synchronization and machine-to-machine communication. DistBelief uses different optimization methods, namely SGD and Sandblaster \cite{dean2012large} . Last, but not least, there are network architectures such as highway networks, residual networks and dense nets  \cite{DBLP:journals/corr/HeZRS15,DBLP:journals/corr/SrivastavaGS15,DBLP:journals/corr/GreffSS16,DBLP:journals/corr/ZillySKS16,DBLP:journals/corr/SrivastavaGS15a} . For example, highway networks are based on LSTM recurrent networks and they allow for the efficient training of deep networks with hundreds of layers based on gradient descent  \cite{DBLP:journals/corr/SrivastavaGS15,DBLP:journals/corr/GreffSS16,DBLP:journals/corr/ZillySKS16} .

\subsubsection{Tools}
\label{subsubsec:Tools}
Tools are also a large factor in DL research and development. Wan et al. observe that DL is at the intersection of NNs, graphical modeling, optimization, pattern recognition and signal processing \cite{Wan2014Deep} , which means there is a fairly high background level required for this area. Good DL tools allow researchers and students to try some basic architectures and create new ones more efficiently. 

Table \ref{table:DLTools} lists some popular DL toolkits and links to the code. Herein, we review some of the DL tools, and the tool analysis below are based on our experiences with these tools. We thank our graduate students for providing detailed feedback on these tools. 

AlexNet \cite{krizhevsky2012imagenet} was a revolutionary paper that re-introduced the world to the results that DL can offer. AlexNet utilizes ReLU because it is several times faster to evaluate than the hyperbolic tangent. AlexNet revealed the importance of pre-processing by incorporating some data augmentation techniques and was able to combat overfitting by using max pooling and dropout layers.  

Caffe \cite{Jia2014Caffe} was the first widely used deep learning toolkit. Caffe is C++ based and can be compiled on various devices, and offers command line, Python, and Matlab interfaces. There are many useful examples provided. The cons of Caffe are that is is relatively hard to install, due to lack of documentation and not being developed by an organized company. For those interested in something other than image processing, (e.g. image classification, image segmentation), it is not really suitable for other areas, such as audio signal processing.

TensorFlow \cite{abadi2016tensorflow} is arguably the most popular DL tool available. It's pros are that TensorFlow (1) is relatively easy to install both with CPU and GPU version on Ubuntu (The GPU version needs CUDA and cuDNN to be installed ahead of time, which is a little complicated); (2) has most of the state-of-the-art models implemented, and while some original implementation are not implemented in TensorFlow, but it is relatively easy to find a re-implementation in TensorFlow; (3) has very good documentation and regular updates; (4) supports both Python and C++ interfaces; and (5) is relatively easy to expand to other areas besides image processing, as long as you understand the tensor processing. One con of TensorFlow is that it is really restricted to Linux applications, as the windows version is barely usable.

MatConvNet \cite{vedaldi2015matconvnet} is a convenient tool, with unique abstract implementations for those very comfortable with using Matlab. It offers many popular trained CNN’s, and the data sets used to train them. It is fairly easy to install. Once the GPU setup is ready (installation of drivers + CUDA support), training with the GPU is very simple. It also offers Windows support. The cons are (1) there is a substantially smaller online community compared to TensorFlow and Caffe, (2) code documentation is not very detailed and in general does not have good online tutorials besides the manual. Lack of “getting started” help besides a very simple example, and (3) GPU setup can be quite tedious. For Windows, Visual Studio is required, due to restrictions on Matlab and its mex setup, as well as Nvidia drivers and CUDA support. On Linux, one has much more freedom, but must be willing to adapt to manual installations of Nvidia drivers, CUDA-support, and more.  
%
%

\begin{table}[ht]
\centering
\caption{Some popular DL tools.}
\label{table:DLTools}

\newcolumntype{L}[1]{>{\raggedright\arraybackslash}p{#1}}
\newcolumntype{R}[1]{>{\raggedright\arraybackslash}p{#1}}

\begin{tabular}{|L{2.75cm}|R{13.25cm}|}

\hline
\textbf{Tool \& Citation}         
& \textbf{Tool Summary and Website} \\ 
\hline
\hline

\multirow{2}{*}{AlexNet \cite{krizhevsky2012imagenet}} 
& A large-scale CNN with a non-saturating,neurons and a very efficient GPU parallel implementation of the convolution operation to make training faster. \\
& Website: \url{http://code.google.com/p/cuda-convnet/} \\ \hline

\multirow{2}{*}{Caffe \cite{Jia2014Caffe}} 
& C++ library with Python and Matlab interfaces. \\
& Website: \url{http://caffe.berkeleyvision.org/} \\ \hline

\multirow{2}{*}{cuda-convnet2 \cite{krizhevsky2012imagenet}}
& The DL tool cuda-convnet2 is a fast C++/CUDA CNN implementation, and can also model any directed acyclic graphs. Training is performed using back-propagation. Offers faster training on Kepler-generation GPUs and multi-GPU training support. \\ 
& Website: \url{https://code.google.com/p/cuda-convnet2/}  \\ \hline

\multirow{2}{*}{gvnn \cite{Handa2016Gvnn}} 
& The DL package gvnn is a NN library in Torch aimed towards bridging the gap between classic geometric computer vision and DL. This DL package is used for recognition, end-to-end visual odometry, depth estimation, etc. \\
& Website: \url{https://github.com/ankurhanda/gvnn}  \\ \hline

\multirow{2}{*}{Keras \cite{chollet2015keras}} 
& Keras is a high-level Python NN library capable of running on top of either TensorFlow or Theano and was developed with a focus on enabling fast experimentation. Keras (1) allows for easy and fast prototyping, (2) supports both convolutional networks and recurrent networks, (3) supports arbitrary connectivity schemes, and (4) runs seamlessly on CPUs and GPUs. \\
& Website: \url{https://keras.io/} and \url{https://github.com/fchollet/keras} \\ \hline

\multirow{2}{*}{MatConvNet \cite{vedaldi2015matconvnet}} 
& A Matlab toolbox implementing CNNs with many pre-trained CNNs for image classification, segmentation, etc. \\
& Website: \url{http://www.vlfeat.org/matconvnet/} \\ \hline

\multirow{2}{*}{MXNet \cite{Chen2015MXNet}} 
& MXNet is a DL library. Features include declarative symbolic expression with imperative tensor computation and differentiation to derive gradients. MXNet runs on  mobile devices to distributed GPU clusters. \\ 
& Website: \url{https://github.com/dmlc/mxnet/} \\ \hline

\multirow{2}{*}{TensorFlow \cite{abadi2016tensorflow}} 
& An open source software library for tensor data flow graph computation. The flexible architecture allows you to deploy computation to one or more CPUs or GPUs in a desktop, server, or mobile devices. \\
& Website: \url{https://www.tensorflow.org/} \\ \hline

\multirow{2}{*}{Theano \cite{AlRfou2016theano}} 
& A Python library that allows you to define, optimize, and efficiently evaluate mathematical expressions involving multi-dimensional arrays. Theano features (1) tight integration with NumPy, (2) transparent use of a GPU, (3) efficient symbolic differentiation, and (4) dynamic C code generation. \\
& Website: \url{http://deeplearning.net/software/theano}  \\ \hline

\multirow{2}{*}{Torch \cite{xxx2017Deep}} 
& Torch is an embeddable scientific computing framework with GPU optimizations, which uses the LuaJIT scripting language and a C/CUDA implementation. Torch includes (1) optimized linear algebra and numeric routines, (2) neural network and energy-based models, and (3) GPU support. \\ 
& Website: \url{http://torch.ch/}  \\ \hline

\end{tabular}

\end{table}

\subsection{DL in other domains}
DL has been utilized in other areas than RS, namely human behavior analysis \cite{baccouche2011sequential,shou2017cdc,tran2015learning,wang2016temporal} , speech recognition \cite{dahl2012context, hinton2012deepneural,graves2013speech} , stereo vision \cite{luo2016efficient} , robotics \cite{levine2016learning} , signal-to-text \cite{venugopalan2015sequence, collobert2011deep, venugopalan2014translating, TanC16, karpathy2015deep} , physics \cite{baldi2014searching, wu2015galileo} , cancer detection \cite{cruz2013deep, fakoor2013using, sirinukunwattana2016locality} , time-series analysis \cite{langkvist2014review, sarkar2015early, kuremoto2014time} , image synthesis \cite{dosovitskiy2015learning,zhao2016energy,reed2016learning,berthelot2017began,TanCAT17,gregor2015draw, radford2015unsupervised} , stock market analysis \cite{ding2015deepStock} , and security applications \cite{yuan2014droid} . These diverse set of applications show the power of DL.

%
%
\section{DL approaches in RS}
\label{sec:DL_Approaches_RS}

\ParSection{Table of remote sensing areas (JOHN-DONE)}
There are many RS tasks that utilize RS data, including automated target detection, pansharpening, land cover and land use classification, time series analysis, change detection, etc. Many of these tasks utilize shape analysis, object recognition, dimensionality reduction, image enhancement, and other techniques, which are all amenable to DL approaches. Table \ref{table:DL_papers} groups DL papers reviewed in this paper into these basic categories. From the table, it can be seen that there is a large diversity of applications, indicating that RS researchers have seen value in using DL methods in many different areas. Several representative papers are reviewed and discussed.

%
%
%
%

\begin{table}[ht]
\caption{DL paper subject areas in remote sensing.} 
\label{table:DL_papers}
\begin{center}       
\begin{tabular}{|l|l|l|l|} 
\hline
\rule[-1ex]{0pt}{3.5ex}  \textbf{Area} & \textbf{References} & \textbf{Area} & \textbf{References} \\
\hline
\hline

\rule[-1ex]{0pt}{3.5ex}  
3D (depth and shape) analysis & 
\cite{Cadena2016, Feng2016, Haque2016, HegdeStanford2016, Huang2016, Kehl2016, Li2015Beyond, Sedaghat2016, xie2015projective} &
Advanced driver assistance systems & 
\cite{Chen2015, Chen2016, chigorin2013system, Ciresan2012, Zeng2015} \\
\hline

\rule[-1ex]{0pt}{3.5ex}  
Animal detection & 
\cite{Salberg2015} & 
Anomaly detection & 
\cite{Li2017Anomaly} \\
\hline

\rule[-1ex]{0pt}{3.5ex}  
Automated Target Recognition & 
\cite{becker2015deep, bentes2015target, besaw2016detecting, besaw2014deep, du2016sar, chen2014sar, morgan2015deep, ni2013sar, sun2013recognition, wang2015application, Zhang2011Multifeature, zhang2015hierarchical} &
Change detection & 
\cite{alcantarilla2016streetview, Gong2016, Pacifici2007, Stent2015, zhao2014deep} \\
\hline

\rule[-1ex]{0pt}{3.5ex}  
Classification & 
\cite{Basu2015, Bazi2014, Cao2016Deep, Cao2016Graph, chen2014spectral, Cheng2016Scene, DelFrate2007Use, Fang2016Using, fu2016semi, Geng2015SAR, han2017scene, He2016Hyperspectral, Hou2015Polarimetric, Hu2015Deep, iftene2016very, Jia2016Convolutional, kontschieder2015deep, Langkvist2016Classification, Lee2016Contextual, Li2016Active, Li2015Deep, Li2014Classification, li2017hyperspectral, Li2016Hyperspectral, lin2013spectral, Lin2013spectral2, liu2015hyperspectral, Liu2015hyperspectral2, liu2016active, ma2016spectral, ma2016hyperspectral, mei2015infrared, Mei2016Integrating, Merentitis2015Automatic, nogueira2017towards, Pan2017RVCANet, papadomanolaki2016benchmarking, piramanayagam2016classification, qin2017object, Rajan2008Active, Wang2015Semisupervised, Yang2016Two, yu2017convolutional, yue2016deep, yue2015spectral, zeggada2016multilabel, zhang2016scene, zhang2017spectral, Zhao2016Learning, Zhong2016Large, zhong2017satcnn} &
Data fusion & 
\cite{Chen2016DeepFusion} \\
\hline

\rule[-1ex]{0pt}{3.5ex}  Dimensionality reduction & 
\cite{ran2016bands, Zabalza2016Novel} &
Disaster analysis/assessment & 
\cite{liu2016geological} \\
\hline

\rule[-1ex]{0pt}{3.5ex}  Environment and water analysis & 
\cite{chen2012design, Landschutzer2013Neural, shi2016cloud, Shi2015Convolutional} &
Geo-information extraction & 
\cite{Lee2015Predicting} \\
\hline

\rule[-1ex]{0pt}{3.5ex}  Human detection & 
\cite{Kehl2016Deep, Kim2015Human, ouyang2012discriminative, tome2016deep} & Image denoising/enhancement &
\cite{Wei2016Universal, zhang2016systematic} \\
\hline

\rule[-1ex]{0pt}{3.5ex}  Image Registration & 
\cite{Quan2016Using} &
Land cover classification &
\cite{Ghamisi2016SelfImproving, kussul2016deep, li2016stacked, Liu2016Terrain, Makantasis2015Deep} \\
\hline

\rule[-1ex]{0pt}{3.5ex}  Land use/classification & 
\cite{Castelluccio2015Land, Cheng2015Effective, Luus2015Multiview, lv2015urban, Ma2016Semisupervised, Midhun2014Deep, Othman2016Using, Penatti2015Deep, romero2016unsupervised, Sun2016Active, uba2016land} &
Object recognition and detection &
\cite{Alexandre2016, Chen2013Aircraft, Chen20153D, Cheng2016Learning, Dahmane2016Object, diao2015object, Georgakis2016, Maturana20153D, Wang2016Differential, wu2016shape, Zhou2015Object}  \\
\hline

\rule[-1ex]{0pt}{3.5ex}  Object tracking & 
\cite{Ondruska2016Deep, masi2016pansharpening} &
Pansharpening & 
\cite{huang2015new} \\
\hline

\rule[-1ex]{0pt}{3.5ex}  Planetary studies & 
\cite{Palafox2015Automated} &
Plant and agricultural analysis & 
\cite{ghazi2017plant, Guan2015Deep, goel2003classification, Kuwata2015Estimating, rebetez1a2016augmenting, Sladojevic2016Deep } \\
\hline

\rule[-1ex]{0pt}{3.5ex}  Road segmentation/extraction & 
\cite{Levi2015StixelNet, li2016road, Mnih2012Learning, wang2015road, Yu2014AutomatedTree, Yu2015AutomatedManhole, Zhong2016Fully} &
Scene understanding & 
\cite{hadsell2009learning, mou2016spatiotemporal, Yuan2015Scene} \\
\hline

\rule[-1ex]{0pt}{3.5ex}  Semantic segmentation/annotation &
\cite{Couprie2013Indoor, Gong2013Deep, kampffmeyer2016semantic, Kaiser2016Learning, Lagrange2015Benchmarking, marmanis2016deep, Marmanis2016Semantic, Paisitkriangkrai2015Effective, Qu2016Deep, sherrah2016fully, vaduva2012deep, Volpi2016Dense, Zhang2015Cosaliency} & Segmentation & 
\cite{Alam2016CRF, Audebert2016Useful, basaeed2016supervisedHierarchial, basaeed2016supervisedRemote, pal2016dcap, Wang2015Deep} \\
\hline

\rule[-1ex]{0pt}{3.5ex}  Ship classification/detection & 
\cite{schwegmann2016very, Tang2015Compressed, zhanga2016SCNN} & Super-resolution & 
\cite{Cui2014Deep, dong2016image, Ducournau2016Deep, Liebel2016SingleImage} \\
\hline

\rule[-1ex]{0pt}{3.5ex}  Traffic flow analysis & 
\cite{huang2014deep, Lv2015Traffic} &
Underwater detection & 
\cite{Elawady2014Sparse, Qin2016DeepFish, Qin2015Underwater, Williams2016Underwater} \\
\hline

\rule[-1ex]{0pt}{3.5ex}  Urban/building & 
\cite{Alidoost2016Knowledge, Brust2015Efficient, de2015urban, huang2016building, Marmanis2015Deep, saito2015building, Vakalopoulou2015Building, xie2015transfer, Yao2016Classification, zhang2016cnn, zhang2016cnnfunctional} &
Vehicle detection/recognition & 
\cite{Cao2016Robust, chen2014vehicle, Goyal2016Novel, Hu2014Deep, Huang2016Vehicle, Huval2015Empirical, jiang2015deep, konoplich2016application, Krishnan2016Vehicle, Lange2016Online, li20163d, wang2014vehicle, wang2016appearance, wang2015night} \\
\hline 

\rule[-1ex]{0pt}{3.5ex}  Weather forecasting & 
\cite{firth2016novel, kovordanyi2009cyclone, yang2016improved} & & \\
\hline 

\end{tabular}
\end{center}
\end{table} 

\ParSection{Summarize some of the approaches. (JOHN-DONE)} 
Due to the large number of recent RS papers, we can't review all of the papers utilizing DL or FL in RS applications. Instead, herein we focus on several papers in different areas of interest that offer creative solutions to problems encountered in DL and FL and should also have a wide interest to the readers. We do provide a summary of all of the DL in RS papers we reviewed online at \url{http://www.cs-chan.com/source/FADL/Online_Paper_Summary_Table.pdf}.

\subsection{Classification}


Classification is the task of labeling pixels (or regions in an image) into one of several classes. The DL methods outlined below utilize many forms of DL to learn features from the data itself and perform classification at state-of-the-art levels. The following discusses classification in HSI, 3D, satellite imagery, traffic sign detection and Synthetic Aperture Radar (SAR).

\textbf{HSI:} HSI data classification is of major importance to RS applications, so many of the DL results we reviewed were on HSI classification. HSI processing has many challenges, including high data dimensionality and usually low numbers of training samples. Chen et al. \cite{Chen2015SpectralSpatial} propose an DBN-based HSI classification framework. The input data is converted to a 1D vector and processed via a DBN with three RBM layers, and the class labels are output from a two-layer logistic regression NN. A spatial classifier using Principal Component Analysis (PCA) on the spectral dimension followed by 1D flattening of a 3D box, a three-level DBN and two level logistic regression classifier. A third architecture uses a combinations of the 1D spectrum and the spatial classifier architecture. He et al. \cite{He2016Hyperspectral} developed a DBN for HSI classification that does not require stochastic gradient descent (SGD) training. Nonlinear layers in the DBN allow for the nonlinear nature of HSI data and a logistic regression classifier is used to classify the outputs of the DBN layers. A parametric depth study showed depth of nine layers produced the best results of depths of 1 to 15, and after a depth of nine, no improvement resulted by adding more layers.

Some of the HSI DL approaches utilize both spectral and spatial information. Ma et al. \cite{ma2016spectral} created a HSI spatial updated deep AE which integrates spatial information. Small training sets are mitigated by a collaborative, representation-based classifier and salt-and-pepper noise is mitigated by a graph-cut-based spatial regularization. Their method is more efficient than comparable kernel-based methods, and the collaborative representation-based classification makes their system relatively robust to small training sets. Yang et al. \cite{Yang2016Two} use a two-channel CNN to jointly learn spectral and spatial features. Transfer learning is used when the number of training samples is limited, where low-level and mid-level features are transferred from other scenes. The network has a spectral CNN and a spatial CNN, and the results are combined in three FC layers. A softmax classifier produces the final class labels. Pan et al. \cite{Pan2017RVCANet} proposed the so called rolling guidance filter and vertex component analysis network (R-VCANet), which also attempts to solve the common problem of lack of HSI training data. The network combines spectral and spatial information. The rolling guidance filter is an edge-preserving filter used to remove noise and small details from imagery. The VCANet is a combination of vertex component analysis \cite{nascimento2005vertex} , which is used to extract pure endmembers, and PCANet \cite{Chan2015PCANet} . A parameter analysis of the number of training samples, rolling times, and the number and size of the convolution kernels. The system performs well even when the training ratio is only 4\%. Lee et al. \cite{Lee2016Contextual} designed a contextual deep fully convolutional DL network with fourteen layers that jointly exploits spatial and HSI spectral features. Variable size convolutional features are utilized to create a spectral-spatial feature map. A novel feature of the architecture is the initial layers uses both $[3 \times 3 \times B]$ convolutional masks to learn spatial features, and $[1 \times 1 \times B]$ for spectral features, where $B$ is the number of spectral bands. The system is trained with a very small number of training samples (200/class).

\textbf{3D:} In 3D analysis, there are several interesting DL approaches. Chen et al. \cite{Chen2016Deep} used a 3D CNN-based feature extraction model with regularization to extract effective spectral-spatial features from HSI. $L_{2}$ regularization and dropout are used to help prevent overfitting. In addition, a virtual enhanced method imputes training samples. Three different CNN architectures are examined: (1) a 1D using only spectral information, consisting of convolution, pooling, convolution, pooling, stacking and logistic regression; (2) a 2D CNN with spatial features, with 2D convolution, pooling, 2D convolution, pooling, stacking, and logistic regression; (3) 3D convolution (2D for spatial and third dimension is spectral); the organization is same as 2D case except with 3D convolution. The 3D CNN achieves near-perfect classification on the data sets. 

Chen et al. \cite{Chen2016DeepFusion} propose a novel 3D CNN to extract the spectral-spatial features of HSI data, a deep 2D CNN to extract the elevation features of Light Detection and Ranging (LiDAR) data, and then a FC DNN to fuse the 2D and 3D CNN outputs. The HSI data are processed via two layers of 3D convolution followed by pooling. The LiDAR elevation data are processed via two layers of 2D convolution followed by pooling. The results are stacked and processed by a FC layer followed by a logistic regression layer. 

Cheng et al. \cite{Cheng2016Learning} developed a rotation-invariant CNN (RICNN), which is trained by optimizing a objective function with a regularization constraint that explicitly enforces the training feature representations before and after rotating to be mapped close to each other. New training samples are imputed by rotating the original samples by $k$ rotation angles. The system is based on AlexNet \cite{krizhevsky2012imagenet}, which has five convolutional layers followed by three FC layers. The AlexNet architecture is modified by adding a rotation-invariant layer that used the output of AlexNet's FC7 layer, and replacing the 1000-way softmax classification layer with a $(C+1)$-layer softmax classifier layer. AlexNet is pretrained, then fine tuned using the small number of HSI training samples. Haque et al. \cite{Haque2016} developed a attention-based human body detector that leverages 4D spatio-temporal signatures and detects humans in the dark (depth images with no RGB content). Their DL system extracts voxels then encodes data using a CNN, followed by a LSTM. An action network gives the class label and a location network selects the next glimpse location. The process repeats at the next time step.

\textbf{Traffic Sign Recognition:} In the area of traffic sign recognition, a nice result came from Ciresan et al. \cite{Ciresan2012} , who created a biologically plausible DNN is based on the feline visual cortex. The network is composed of multiple columns of DNNs, coded for parallel GPU speedup. The output of the columns is averaged. It outperforms humans by a factor of two in traffic sign recognition.

\textbf{Satellite Imagery:} In the area of satellite imagery analysis, Zhang et al. \cite{zhang2016scene} propose a gradient-boosting random convolutional network (GBRCN) to classify very high resolution (VHR) satellite imagery. In GBRCN, a sum of functions (called boosts) are optimized. A modified multi-class softmax function is used for optimization, making the optimization task easier. SGD is used for optimization. Proposed future work was to utilize a variant of this method on HSI. Zhong et al. \cite{zhong2017satcnn} use efficient small CNN kernels and a deep architecture to learn hierarchical spatial relationships in satellite imagery. A softmax classifier output class labels based on the CNN DL outputs. The CPU handles preprocessing (data splitting and normalization), while the GPU performs convolution, ReLU and pooling operations, and the the CPU handles dropout and softmax classification. Networks with one to three convolution layers are analyzed, with receptive fields of $10 \times 10$ to $1000 \times 1000$. SGD is used for optimization. A hyper-parameter analysis of the learning rate, momentum, training-to-test ratio, and number of kernels in the first convolutional layer were also performed. 

\textbf{SAR:} In the area of SAR processing, De et al. \cite{de2015urban} use DL to classify urban areas, even when rotated. Rotated urban target exhibit different scattering mechanisms, and the network learns the $\alpha$ and $\gamma$ parameters from the HH, VV and HV bands (H=Horizontal, V-Vertical polarization). Bentes et al. \cite{bentes2015target} use a constant false alarm rate (CFAR) processor on SAR data followed by $N$ AEs. The final layer associates the learned features with class labels. Geng et al. \cite{Geng2015SAR} used a eight-layer network with a convolutional layer to extract texture features from SAR imagery, a scale transformation layer to aggregate neighbor features, four Stacked AE (SAE) layers for feature optimization and classification, and a two-layer post processor. Gray level co-occurrence matrix and Gabor features are also extracted, and average pooling is used in layer two to mitigate noise.

\subsection{Transfer Learning} 

Transfer learning utilizes training in one image (or domain) to enable better results in another image (or domain). If the learning crosses domains, then it may be possible to utilize lower to mid-level features learned from on domain in the other domain. 

Marmanis et al. \cite{marmanis2016deep} attacked the common problem in RS of limited training data by utilizing transfer learning across domains. They utilized a CNN pretrained on the ImageNet dataset, and extracted an initial set of representations from orthoimagery. These representations are then transferred to a CNN classifier. This paper developed a novel cross-domain feature fusion system. Their system has seven convolution layers followed by two long MLP layers, three convolution layers, two more large MLP layers, and finally a softmax classifier. They extract feature from the last layer, since the work of Donahue et al. \cite{Donahue2014DeCAF} showed that most of the discriminative information is contained in the deeper layers. In addition, they take features from the large ($1 \times 1 \times 4096$) MLP, which is a very long vector output, and transform it into a 2D array followed by a large convolution ($91 \times $91) mask layer. This is done because the large feature vector is a computational bottleneck, while the 2D data can very effectively be processed via a second CNN. This approach will work if the second CNN can learn (disentangle) the information in the 2D representation through its layers. This approach is very unique and it raises some interesting questions about alternate DL architectures. This approach was also successful because the features learned by the original CNN were effective in the new image domain. 

Penatti et al. \cite{Penatti2015Deep} asked if deep features generalize from everyday objects to remote sensing and aerial scene domains? A CNN was trained for recognizing everyday objects using ImageNet. The CNNs analyzed performed well, in areas well outside of their training. In a similar vein, Salberg \cite{Salberg2015} use CNNs pretrained on ImageNet to detect seal pups in aerial RS imagery. A linear SVM was used for classification. The system was able to detect seals with high accuracy.

\subsection{3D Processing and Depth Estimation}

Cadena et al. \cite{Cadena2016} utilized multi-modal AEs for RGB imagery, depth images, and semantic labels. Through the AE, the system learns a shared representation of the distinct inputs. The AEs first denoise the given inputs. Depth information is processed as inverse depth (so sky can be handled). Three different architectures are investigated. Their system was able to make a sparse depth map more dense by fusing RGB data.

Feng et al. \cite{Feng2016} developed a content-based 3D shape retrieval system. The system uses a low-cost 3D sensor (e.g. Kinect or Realsense) and a database of 3D objects. An ensemble of AEs learns compressed representations of the 3D objects, and the AE act as probabilistic models which output a likelihood score. A domain adaptation layer uses weakly supervised learning to learn cross-domain representations (noisy imagery and 3D computer aided design (CAD)). The system uses the AE encoded objects to reconstruct the objects, and then additional layers rank the outputs based on similarity scores. Segaghat et al. \cite{Sedaghat2016} use a 3D voxel net that predicts the object pose as well as its class label, since 3D objects can appear very differently based on their poses. The results were tested on LiDAR data, CAD models, and RGB plus depth (RGBD) imagery. Finally, Zelener et al. \cite{zelener2016cnn} labels missing 3D LiDAR points to enable the CNN to have higher accuracy. A major contribution of this method is creating normalized patches of low-level features from the 3D LiDAR point cloud. The LiDAR data is divided into multiple scan lines, and positive and negative samples. Patches are randomly selected for training. A sliding block scheme is used to classify the entire image.

\subsection{Segmentation}
Segmentation means to process imagery and divide it into regions (segments) based on the content.
Basaeed et al. \cite{basaeed2016supervisedHierarchial} use a committee of CNNs that perform multi-scale analysis on each band to estimate region boundary confidence maps, which are then inter-fused to produce an overall confidence map. A morphological scheme integrates these maps into a hierarchical segmentation map for the satellite imagery.

Couprie et al. \cite{Couprie2013Indoor} utilized a multi-scale CNN to learn features directly from RGBD imagery. The image RGB channels and the depth image are transformed through a Laplacian pyramid approach, where each scale is fed to a 3-stage convolutional network that create feature maps. The feature maps of all scales are concatenated (the coarser-scale feature maps are upsampled to match the size of the finest-scale map). A parallel segmentation of the image into superpixels is computed to exploit the natural contours of the image. The final labeling is obtained by the aggregation of the classifier predictions into the superpixels.

In his Master's thesis, Kaiser \cite{Kaiser2016Learning} (1) generated new ground truth datasets for three different cities consisting of VHR aerial images with ground sampling distance on the order of centimeters and corresponding pixel-wise object labels, (2) developed FC networks (FCNs) were used to perform pixel-dense semantic segmentation, (3) created two modifications of the FCN architecture were found that gave performance improvements, and (4) utilized transfer learning was shown using FCN model was trained on huge and diverse ground truth data of the three cities, which achieved good semantic segmentations of areas not used for training.  

L{\"{a}}ngkvist et al. \cite{Langkvist2016Classification} applied a CNN to orthorectified multispectral imagery (MSI) and a digital surface model of a small city for a full, fast and accurate per-pixel classification. The predicted low-level pixel classes are then used to improve the high-level segmentation. Various design choices of the CNN architecture are evaluated and analyzed.

\subsection{Object Detection and tracking}


Object detection and tracking is important in many RS applications. It requires understanding at a higher level than just at the pixel-level. Tracking then takes the process one step further and estimates the location of the object over time.

Diao et al. \cite{diao2015object} propose a pixel-wise DBN for object recognition. A sparse RBM is trained in an unsupervised manner. Several layers of RBM are stacked to generate a DBN. For fine-tuning, a supervised layer is attached to the top of the DBN and the network is trained using BP with a sparse penalty constraint. Ondruska et al. \cite{Ondruska2016Deep} used RNN to track multiple objects from 2D laser data. This system uses no hand-coded plant or sensor models (these are required in Kalman filters). Their system uses an end-to-end RNN approach that maps raw sensor data to a hidden sensor space. The system then predicts the unoccluded state from the sensor space data. The system learns directly from the data and does not require a plant or sensor model.

Schwegmann et al. \cite{schwegmann2016very} use a very deep Highway Network for ship discrimination in SAR imagery, and a three-class SAR dataset is also provided. Deep networks of 2, 20, 50 and 100 layers were tested, and the 20 layer network had the best performance. Tang et al. \cite{Tang2015Compressed} utilized a hybrid approach in both feature extraction and machine learning. For feature extraction, the Discrete Wavelet Transform (DWT) LL, LH, HL and HH (L=Low Frequency, H = High Frequency) features from the JPEG2000 CDF9/7 encoder were utilized. The LL features were inputs to a Stacked DAE (SDAE). The high frequency DWT subbands LH, HL and HH are inputs to a second SDAE. Thus the hand-coded wavelets provide features, while the two SDAEs learn features from the wavelet data. After initial segmentation, the segmentation area, major-to-minor axis ratio and compactness, which are classical machine learning features, are also used to reduce false positives. The training data are normalized to zero mean and unity variance, and the wavelet features are normalized to the $[0,1]$ range. The training batches have different class mixtures, and 20\% of inputs are dropped to the SDAEs and there is a 50\% dropout in the hidden units. The extreme learning machine is used to fuse the low-frequency and high-frequency subbands. An online-sequential extreme learning machine, which is a feedforward shallow NN, is used for classification. 

Two of the most interesting results were developed to handle incomplete training data, and how object detectors emerge from CNN scene classifiers. Mnih et al. \cite{Mnih2012Learning} developed two robust loss functions to deal with incomplete training labeling and misregistration (location of object in map) is inaccurate. A NN is used to model pixel distributions (assuming they are independent). Optimization is performed using expectation maximization. Zhou et al. \cite{Zhou2015Object} show that object detectors emerge from CNNs trained to perform scene classification. They demonstrated that the same CNN can perform both scene recognition and object localization in a single forward pass, without having to explicitly learn the notion of objects. Images had their edges removed such that each edge removal produces the smallest change to the classification discriminant function. This process is repeated until the image is misclassified. The final product of that analysis is a set of simplified images which still have high classification accuracies. For instance, in bedroom scenes, 87\% of these contained a bed. To estimate the empirical receptive field (RF), the images were replicated and random $11 \times 11$ occluded patches were overlaid. Each occluded image is input to the trained DL network and the activation function changes are observed; a large discrepancy indicates the patch was important to the classification task. From this analysis, a discrepancy map is built for each image. As the layers get deeper in the network, the RF size gradually increases and the activation regions are semantically meaningful. Finally, the objects that emerging in one specific layer indicated that the network was learning object categories (dogs, humans, etc.) This work indicates there is still extensive research to be performed in this area.

\subsection{Super-resolution}

Super-resolution analysis attempts to infer sub-pixel information from the data. Dong et al. \cite{dong2016image} utilized a DL network that learns a mapping between the low and high-resolution images. The CNN takes the low-resolution (LR) image as input and outputs the high-resolution (HR) image. In this method, all layers of the DL system are jointly optimized. In a typical super-resolution pipeline with sparse dictionary learning, image patches are densely sampled from the image and encoded in a sparse dictionary. The DL system does not explicitly learn the sparse dictionaries or manifolds for modeling the image patches. The proposed system provides better results than traditional methods and has a fast on-line implementation. The results improve when more data is available or when deeper networks are utilized.

\subsection{Weather Forecasting}

Weather forecasting attempts to use physical laws combined with atmospheric measurements to predict weather patterns, precipitation, etc. The weather effects virtually every person on the planet, so it is natural that there are several RS papers utilizing DL to improve weather forecasting. DL ability to learn from data and understand highly-nonlinear behavior shows much promise in this area of RS.

Chen et al. \cite{chen2012design} utilize DBNs for drought prediction. A three-step process (1) computes the Standardized Precipitation Index (SPI), which is effectively a probability of precipitation, (2) normalizes the SPI, and (3) determines the optimal network architecture (number of hidden layers) experimentally. Firth \cite{firth2016novel} introduced a Differential Integration Time Step network composed of a traditional NN and a weighted summation layer to produce weather predictions. The NN computes the derivatives of the inputs. These elemental building blocks are used to model the various equations that govern weather. Using time series data, forecast convolutions feed time derivative networks which perform time integration. The output images are then fed back to the inputs at the next time step. The recurrent deep network can be unrolled. The network is trained using backpropagation. A pipelined, parallel version is also developed for efficient computation. The model outperformed standard models. The model is efficient and works on a regional level, versus previous models which are constrained to local levels.

Kovordanyi et al. \cite{kovordanyi2009cyclone} utilized NNs in cyclone track forecasting. The system uses a multi-layer NN designed to mimic portions of the human visual system to analyze National Oceanic and Atmospheric Administration's Advanced Very High Resolution Radiometer (NOAA AVHRR) imagery. At the first network level, shape recognition focuses on narrow spatial regions, e.g. detecting small cloud segments. Regions in the image can be processed in parallel using a matrix feature detector architecture. Rotational variations, which are paramount in cyclone analysis, are incorporated into the architecture. Later stages combine previous activations to learn more complex and larger structures from the imagery. The output at the end of the processing system is a directional estimator of cyclone motion. The simulation tool Leabra++ (\url{http://ccnbook.colorado.edu/}) was used. This tool is designed for simulating brain-like artificial NNs (ANNs). There are a total of five layers in the system: an input layer, three processing layers, and an output layer. During training, images were divided into smaller blocks and rotated, shifted, and enlarged. During training, the network was first given inputs and allowed to settle to steady state. Weak activations were suppressed, with at most $k$ nodes were allowed to stay active. Then the inputs and correct outputs were presented to the network and the weights are all zeroed. The learned weights are a combination of the two schemes. Conditional PCA and contrastive Hebbian learning were used to train the network. The system was very effective if the Cyclone's center varied about 6\% or less of the original image size, and less effective if there was more variation.

Shi et al. \cite{Shi2015Convolutional} extended the FC LSTM (FC-LSTM) network that they call ConvLSTM, which has convolutional structures in the input-to-state and state-to-state transitions. The application is precipitation nowcasting, which takes weather data and predicts immediate future precipitation. ConvLSTM used 3D tensors whose last two dimensions are spatial to encode spatial data into the system. An encoding LSTM compresses the input sequence into a latent tensor, while the forecasting LSTM provides the predictions.

\subsection{Automated object and target detection and identification}

Automated object and atutomated target detection and identification is an important RS task for military applications, border security, intrusion detection, advanced driver assistance systems, etc. Both automated target detection and identification are hard tasks, because usually there are very few training samples for the target (but almost all samples of the training data are available as non-target), and often there are large variations in aspect angles, lighting, etc.

Ghazi et al. \cite{ghazi2017plant} used DL to identify plants in photographs using transfer parameter optimization. The main contributions of this work are (1) a state-of-the-art plant detection transfer learning system, and (2) an extensive study of fine-tuning, iteration size, batch size and data augmentation (rotation, translation, reflection, and scaling). It was found that transfer learning (and fine tuning) provided better results than training from scratch. Also, if training from scratch, smaller networks performed better, probably due to smaller training data. The authors suggest using smaller networks in these cases. Performance was also directly related to the network depth. By varying the iteration sizes, it is seen that the validation accuracies rise quickly initially, then grow slowly. The networks studied are all resilient to overfitting. The batch sizes were varied, and larger batch sizes resulted in higher performance at the expense of longer training times. Data augmentation also had a significant effect on performance. The number of iterations had the most effect on the output, followed by the number of patches, and the batch size had the least significant effect. There were significant differences in training times of the systems. Li et al. \cite{Li2017Anomaly} used DL for anomaly detection. In this work, a reference image with pixel pairs (a pair of samples from the same class, and a pair from different classes) is required. By using transfer learning, the system is utilized on another image from the same sensor. Using vicinal pixels, the algorithm recognizes central pixels as anomalies. A 16-level network contains layers of convolution followed by ReLUs. A fully-connected layer then provides output labels.

\subsection{Image Enhancement}

Image enhancement includes many areas such as pansharpening, denoising, image registration, etc. Image enhancement is often performed prior to feature extraction or other image processing steps. Huang et al. \cite{huang2015new} utilize a modified sparse denoising AE (SPDAE), denoted MSDA, which uses the SPDAE to represent the relationship between the HR image patches as clean data to the lower spatial resolution, high spectral resolution MSI image as corrupted data. The reconstruction error drives the cost function and layer-by-layer training is utilized. Quan et al. \cite{Quan2016Using} use DL for SAR image registration, which is in general a harder problem than RGB image registration due to high speckle noise. The RBM learns features useful for image registration, and the random sample consensus (RANSAC) algorithm is run multiple times to reduce outlier points. 

Wei et al. \cite{Wei2016Universal} applied a five-layer DL network to perform image quality improvement. In their approach, degraded images are modeled as downsampled images that are degraded by a blurring function and additive noise. Instead of trying to estimate the inverse function, a DL network performs feature extraction at layer 1, then the second layer learns a matrix of kernels and biases to perform non-linear operations to layer 1 outputs. Layers 3 and 4 repeat the operations of layers 1 and 2. Finally, an output layer reconstructs the enhanced imagery. They demonstrated results with non-uniform haze removal and random amounts of Gaussian noise. Zhang et al. \cite{zhang2016systematic} applied DL to enhance thermal imagery, based on first compensating for the camera transfer function (small-scale and large-scale nonlinearities), and then super-resolution target signature enhancement via DL. Patches are extracted from low-resolution imagery, and the DL learns feature maps from this imagery. A nonlinear mapping of these feature maps to a HR image are then learned. SGD is utilized to train the network.

\subsection{Change Detection}

Change detection is the process of utilizing two registered RS images taken at different times and detecting the changes, which can be due to natural phenomenon such as drought or flooding, or due to man-made phenomenon, such as adding a new road or tearing down an old building. We note that there is a paucity of DL research into change detection. Pacifici et al. \cite{Pacifici2007} used DL for change detection in VHR satellite imagery. The DL system exploits the multispectral and multitemporal nature of the imagery. Saturation is avoided by normalizing data to $[-1,1]$ range. To mitigate illumination changes, band ratios such as blue/green are utilized. These images are classified according to (1) man-made surfaces, (2) green vegetation, (3) bare soil and dry vegetation, and (4) water. Each image undergoes a classification and a multitemporal operator creates a change mask. The two classification maps and the change mask are fused using an AND operator.

\subsection{Semantic Labeling}

Semantic labeling attempts to label scenes or objects semantically, such as ``there is a truck next to the tree''. Sherrah et al. \cite{sherrah2016fully} utilized the recent development of FC NNs (FC-CNNs), which were developed by Long et al. \cite{long2015fully}. The FC-CNN is applied to remote sensed VHR imagery. In their network, there is no downsampling. The system labels images semantically pixel-by-pixel. Xie et al. \cite{xie2015transfer} used transfer learning to avoid training issues due to scarce training data, transfer learning is utilized. A FC CNN trains in daytime imagery and predicts nighttime lights. The system also can infer poverty data from the night lights, as well as delineating man-made structures such as roads, buildings and farmlands. The CNN was trained on ImageNet and uses the NOAA nighttime remote sensing satellite imagery. Poverty data was derived from a living standards measurement survey in Uganda. Mini-batch gradient descent with momentum, random mirroring for data augmentation, and 50\% dropout was used to help avoid overfitting. The transfer learning approach gave higher performance in accuracy, F1 scores, precision and area under the curve. 

\subsection{Dimensionality reduction}
\ParSection{(JOHN-DONE)}

HSI are inherently highly dimensional, and often contain highly correlated data. Dimensionality reduction can significantly improve results in HSI processing. Ran et al. \cite{ran2016bands} split the spectrum into groups based on correlation, then apply $m$ CNNs in parallel, one for each band group. The CNN output are concatenated and then classified via a two-layer FC-CNN. Zabalza et al. \cite{Zabalza2016Novel} used segmented SAEs are utilized for dimensionality reduction. The spectral data are segmented into $k$ regions, each of which has a SAE to reduce dimensionality. Then the features are concatenated into a reduced profile vector. The segmented regions are determine by using the correlation matrix of the spectrum. In Ball et al. \cite{ball2014hyperspectral} , it was shown that band selection is task and data dependent, and often better results can be found by fusing similarity measures versus using correlation, so both of these methods could be improved using similar approaches. Dimensionality reduction is an important processing step in many classification algorithms \cite{Ball2007LevelSetBBA, Ball2007LevelSetSID} , pixel unmixing \cite{keshava2003survey, keshava2002spectral, anderson2012spectral, winter2000comparison, charles2011learning, romero2014unsupervised, ball2007hyperspectral} , etc.

%
%
\section{Unsolved challenges and opportunities for DL in RS}
\label{sec:ChallengesOpportunitiesDLRS}

\ParSection{Overview (JOHN-DONE)} DL applied to RS has many challenges and open issues. Table \ref{table:DLFLSurveyPapers} gives some representative DL and FL survey papers and discusses their main content. Based on these reviews, and the reviews of many survey papers in RS, we have identified the following major open issues in DL in RS. Herein, we focus on unsolved challenges and opportunities as it relates to 
(i) inadequate data sets (\ref{subsec:ChallengesOpportunities_i}), 
(ii) human-understandable solutions for modelling physical phenomena (\ref{subsec:ChallengesOpportunities_ii}),
(iii) Big Data (\ref{subsec:ChallengesOpportunities_iii}), 
(iv) non-traditional heterogeneous data sources (\ref{subsec:ChallengesOpportunities_iv}), 
(v) DL architectures and learning algorithms for spectral, spatial and temporal data  (\ref{subsec:ChallengesOpportunities_v}), 
(vi) transfer learning (\ref{subsec:ChallengesOpportunities_vi}), 
(vii) an improved theoretical understanding of DL systems (\ref{subsec:ChallengesOpportunities_vii}), 
(viii) high barriers to entry (\ref{subsec:ChallengesOpportunities_viii}), and
(ix) training and optimizing the DL (\ref{subsec:ChallengesOpportunities_ix}).

%
%
\ParSection{Table of RS survey papers (JOHN-DONE)}
%
%
\begin{table}[ht]
\centering
\caption{Representative DL and FL Survey papers.}
\label{table:DLFLSurveyPapers}

\newcolumntype{L}[1]{>{\raggedright\arraybackslash}p{#1}}
\newcolumntype{R}[1]{>{\raggedright\arraybackslash}p{#1}}

\begin{tabular}{|L{0.8cm}|R{15cm}|}

\hline
\textbf{Ref.}         
& \textbf{Paper Contents} \\ 
\hline
\hline

\cite{Arel2010} 
& A survey paper on DL. Covers CNNs, DBNs, etc. \\
\hline

\cite{atkinson1997introduction} 
& Brief intro to neural networks in remote sensing. \\
\hline

\cite{bengio2012unsupervised} 
& Overview of unsupervised feature learning and deep learning. Provides overview of probabilistic models (undirected graphical, RBM, AE, SAE, DAE, contractive autoencoders, manifold learning, difficulty in training deep networks, handling high-dimensional inputs, evaluating performance, etc.) \\
\hline

\cite{cavallaro2015understanding} 
& Examines big-data impacts on SVM machine learning. \\
\hline

\cite{Cheng} 
& Covers about 170 publications in the area of scene classification and discusses limitations of datasets and problems associated with high-resolution imagery. They discuss limitations of handcrafted features such as texture descriptors, GIST, SIFT, HOG. \\
\hline

\cite{Deng2014} 
& A good overview of architectures, algorithms, and applications for DL. Three important reasons for DL success are (1) GPU units, (2) recent advances in DL research. In addition, we note that (3) would be success of DL in many image processing challenges. DL is at the intersection of machine learning, Neural Networks, optimization, graphical modeling, pattern recognition, probability theory and signal processing. They discuss generative, discriminative, and hybrid deep architectures. They show there is vast room to improve the current optimization techniques in DL. \\
\hline

\cite{Egmont-Petersen2002} 
& Overview of NN in image processing. \\
\hline

\cite{huang2015trends} 
& Discusses trends in extreme learning machines, which are linear, single hidden layer feedforward neural networks. ELMs are comparable or better than SVMs in generalization ability. In some cases, ELMs have comparable performance to DL approaches. They generally have high generalization capability, are universal approximators, don't require iterative learning, and have a unified learning theory. \\
\hline

\cite{jia2013feature} 
& Provides overview of feature reduction in remote sensing imagery. \\
\hline

\cite{liu2016survey} 
& A survey of deep neural networks, including the AE, the CNN, and applications. \\
\hline

\cite{lu2007survey} 
& Survey of image classification methods in remote sensing. \\
\hline

\cite{petersson2016hyperspectral} 
& Short survey of DL in hyperspectral remote sensing. In particular, in one study, there was a definite sweet spot shown in the DL depth. \\
\hline

\cite{plaza2009recent} 
& Overview of shallow HSI processing. \\
\hline

\cite{plaza2004quantitative} 
& Overview of shallow endmember extraction algorithms. \\
\hline

\cite{Schmidhuber2015} 
& An in-depth historical overview of DL. \\
\hline

\cite{wang2015survey} 
& History of DL. \\
\hline

\cite{wang2016review} 
& A review of road extraction from remote sensing imagery. \\
\hline

\cite{Yu2011} 
& A review of DL in signal and image processing. Comparisons are made to shallow learning, and DL advantages are given. \\
\hline

\cite{Zhang2016} 
& Provides a general framework for DL in remote sensing. Covers four RS perspectives: (1) image processing, (2) pixel-based classification, (3) target recognition, and (4) scene understanding. \\
\hline

\end{tabular}
\end{table}

%
%
\subsection{Inadequate data sets}
\label{subsec:ChallengesOpportunities_i}
\textbf{Open Question \#1a: How can DL systems work well with limited datasets?}

\ParSection{Old data sets getting obsolete, new needed(JOHN - DONE)} 
There are two main issues with most current RS data sets. Table \ref{table:HSI_dataset} provides a summary of the more common open-source datasets for the DL papers utilizing HSI data. Many of these papers utilized custom datasets, and these are not reported. Table \ref{table:HSI_dataset} shows that the most commonly used datasets were Indian Pines, Pavia University, Pavia City Center, and Salinas.

\begin{table}[ht]
\caption{HSI Dataset Usage.} 
\label{table:HSI_dataset}
\begin{center}       
\begin{tabular}{|l|c|} 

\hline
\rule[-1ex]{0pt}{3.5ex}  \textbf{Dataset and Reference}         &  \textbf{Number of uses} \\
\hline
\hline

IEEE GRSS 2013 Data Fusion Contest \cite{IEEEGRSSDataFusion2013Dataset}     & 4   \\
\hline

IEEE GRSS 2015 Data Fusion Contest \cite{IEEEGRSSDataFusion2015Dataset}     & 1   \\
\hline

IEEE GRSS 2016 Data Fusion Contest \cite{IEEEGRSSDataFusion2016Dataset}     & 2   \\
\hline

Indian Pines \cite{IndianPinesDataset}	                                    & 27  \\
\hline

Kennedy Space Center \cite{KennedySpaceCenterDataSet}                       & 8   \\
\hline

Pavia City Center \cite{PaviaDatasets}                                      & 13  \\
\hline

Pavia University \cite{PaviaDatasets}                                       & 19  \\ 
\hline

Salinas \cite{SalinasDataset}                                               & 11  \\
\hline

Washington DC Mall \cite{WashingtonDCMallDataset}                            & 2   \\
\hline

\end{tabular}
\end{center}
\end{table}

A very detailed online table (too large to put in this paper) is provided which lists each paper cited in Table \ref{table:DL_papers}. For each paper, a summary of the contributions is given, the datasets utilized are listed, and the papers are categorized in areas (e.g. HSI/MSI, SAR, 3D, etc.). The interested reader can find this at \url{http://www.cs-chan.com/source/FADL/Online_Dataset_Summary_Table.pdf}.

While these are all good datasets, the accuracies from many of the DL papers are nearly saturated. This is shown clearly in Table \ref{table:HSI_OA_Results}. Table \ref{table:HSI_OA_Results} shows results for the HSI DL papers against the commonly-used datasets Indian Pines, Kennedy Space Center, Pavia City Center, Pavia University, Salinas, and the Washington DC Mall. First, OA results must be taken with a grain of salt, since (1) the number of training samples per class can differ for each paper, (2) the number of testing samples can also differ, (3) classes with few relative training samples can even have 0\% overall accuracy, and if there is a large number of test samples of the other classes, the final overall accuracies can still be high. Nevertheless, it is clear from examination of the table that the Indian Pines, Pavia City Center, Pavia University and Salinas datasets are basically saturated. 

In general, it is good to compare new methods to commonly-used datasets, new and challenging datasets are required. Cheng et al. \cite{Cheng, Cheng2016} point out that many existing RS datasets lack image variation, diversity, and have a small number of classes. Datasets are also saturating with accuracy. They created a large-scale benchmark dataset, "NWPU-RESISC45", which attempts to address all of these issues, and made it available to the RS community. The RS community can also benefit from a common practice in the CV community: publishing both datasets and algorithms online, allowing for more comparisons. A typical RS paper may only test their algorithm on two or three images and against only a few other methods. In the CV community, papers usually compare against a large amount of other methods and with many datasets, which may provide more insight about the proposed solution and how it compares to previous work. 

\begin{table}[ht]
\centering

\caption{HSI Overall Accuracy Results in percent. IP = Indian Pines, KSC = Kennedy Space Center, PaCC = Pavia City Center, Pau = Pavia University, Sal = Salinas, DCM = Washington DC Mall. Results higher than 99\% are in bold.}
\label{table:HSI_OA_Results}
\begin{tabular}{|c|c|c|c|c|c|c|}
\hline

Ref     & IP & KSC & PaCC & PaU & Sal & DCM \\ \hline

\cite{Alam2016CRF} & 93.4 &  &  &  &  &  \\ \hline
\cite{Bazi2014} & 98.0 & 98.0 &  & 98.4 &  & 95.4 \\ \hline
\cite{Cao2016Deep} & 97.6 &  &  &  &  &  \\ \hline
\cite{Cao2016Graph} & 97.6 &  &  &  &  &  \\ \hline
\cite{Chen2014Deep} &  & 98.8 & 98.5 &  &  &  \\ \hline
\cite{Chen2015SpectralSpatial} & 96.0 &  & \textbf{99.1} &  &  &  \\ \hline
\cite{fu2016semi} &  &  &  & 94.3 &  &  \\ \hline
\cite{Chen2016DeepFusion} & 89.6 &  &  & 87.1 &  &  \\ \hline
\cite{He2016Hyperspectral} &  & 96.6 &  &  &  &  \\ \hline
\cite{Hu2015Deep} & 90.2 &  &  & 92.6 & 92.6 &  \\ \hline
\cite{Jia2016Convolutional} &  & 84.2 &  &  &  &  \\ \hline
\cite{Lee2016Contextual} & 92.1 &  &  & 94.0 &  &  \\ \hline
\cite{Li2016Active} &  &  &  & \textbf{99.9} &  &  \\ \hline
\cite{Li2015Deep} & 96.3 &  &  &  &  &  \\ \hline
\cite{li2017hyperspectral} & 94.3 &  &  & 96.5 & 94.8 &  \\ \hline
\cite{Li2016Hyperspectral} & 97.6 &  &  & \textbf{99.4} & 98.8 &  \\ \hline
\cite{lin2013spectral} &  & 96.0 & 85.6 &  &  &  \\ \hline
\cite{Liu2015hyperspectral2} & 91.9 &  & \textbf{99.8} & 96.7 & 95.5 &  \\ \hline
\cite{liu2016active} &  &  & 94.0 & 93.5 &  &  \\ \hline
\cite{Ma2016Semisupervised} & 86.5 &  &  & 82.6 &  &  \\ \hline
\cite{ma2016spectral} & \textbf{99.2} &  & \textbf{99.9} &  &  &  \\ \hline
\cite{ma2016hyperspectral} &  &  & 96.0 &  &  & 83.8 \\ \hline
\cite{Makantasis2015Deep} & 98.9 &  & \textbf{99.9} &  & \textbf{99.5} &  \\ \hline
\cite{Mei2016Integrating} & 95.7 &  &  & \textbf{99.6} & 97.4 &  \\ \hline
\cite{Merentitis2015Automatic} & 96.8 &  &  &  &  &  \\ \hline
\cite{Midhun2014Deep} & 79.3 &  &  &  &  &  \\ \hline
\cite{Pan2017RVCANet} & 97.9 & 97.9 &  & 96.8 &  &  \\ \hline
\cite{Rajan2008Active} &  & 80.5 &  &  &  &  \\ \hline
\cite{ran2016bands} & 93.1 &  &  & 95.6 &  &  \\ \hline
\cite{slavkovikj2015hyperspectral} & 96.6 &  &  &  &  &  \\ \hline
\cite{Sun2016Active} & 73.0 & 89.0 &  &  &  &  \\ \hline
\cite{Wang2015Semisupervised} & 93.1 &  &  & 90.4 & \textbf{99.4} &  \\ \hline
\cite{Yang2016Two} & 95.6 &  &  &  &  &  \\ \hline
\cite{yue2016deep} &  &  &  & 67.9 & 85.2 &  \\ \hline
\cite{yue2015spectral} &  &  & 95.2 &  &  &  \\ \hline
\cite{Zabalza2016Novel} & 82.1 &  & 97.4 &  &  &  \\ \hline
\cite{zhang2017spectral} & \textbf{99.7} &  &  & 98.8 &  &  \\ \hline
\cite{Zhao2016Learning} &  &  & \textbf{99.7} & 96.8 &  & \\ \hline                                                      
\end{tabular}
\end{table}

An extensive table of all the datasets utilized in the papers reviewed for this survey paper is made available online, because it is too large to include in this paper. The table is available at \url{http://www.cs-chan.com/source/FADL/Online_Dataset_Summary_Table.pdf}. The table lists the dataset name, briefly describes the datasets, provides a URL (if one is available), and a reference. Hopefully, this table will assist researchers starting work and looking for publicly available datasets.

\textbf{Open Question \#1b: How can DL systems work well with limited training data?}

\ParSection{Small training available}
The second issue is that most RS data has a very small amount of training data available. Ironically, in the CV community, DL has an insatiable hunger for larger and larger data sets (millions or tens of millions of training images), while in the RS field, there is also a large amount of imagery, however, there is usually only a small amount with labeled training samples. RS training data is expensive, error-prone, and usually requires some expert interpretation, which is typically expensive (in terms of time, effort involved, and money) and often requires large amounts of field work and many hours or days post processing the data. Many DL systems, especially those with large numbers of parameters, require large amounts of training data, or else they can easily overtrain and not generalize well. This problem has also plagued other shallow systems as well, such as SVMs.

Approaches used to mitigate small training samples are (1) transfer learning, where one trains on other imagery to obtain low-level to mid-level features which can still be used, or on other images from the same sensor - Transfer learning is discussed in section \ref{subsec:ChallengesOpportunities_vi} below; (2) data augmentation, including affine transformations, rotations, small patch removal, etc.; (3) using ancillary data, such as data from other sensor modalities (e.g. LiDAR, digital elevation models (DEMs), etc.); and (4) unsupervised training, where training labels are not required, e.g. AEs and SAEs. SAEs that have a diabolo shape will force the AE network to learn a lower-dimensional representation.

Ma et al. \cite{ma2016spectral} utilized a DAE and employed a collaborative representation-based classification, where each test sample can be linearly represented by the training samples in the same class with the minimum residual. In classification, features of each sample are approximated with a linear combination of features of all training sample within each class, and the label can be derived according to the class which best approximates the test features. Interested reader is referred to references $46$--$48$ in \cite{ma2016spectral} for more information on collaborative representation. Tao et al. \cite{Tao2015Unsupervised} utilized a Stacked Sparse AE (SSAE) that was shown to be very generalizable and performed well in cases when there were limited training samples. Ghamasi et al. \cite{Ghamisi2016SelfImproving} use Darwinian particle swarm optimization in conjunction with CNNs to select an optimal band set for classifying HSI data. By reducing the input dimensionality, fewer training samples are required. Yang et al. \cite{Yang2016Two} utilized dual CNNs and transfer learning to improve performance. In this method, the lower and middle layers can be trained on other scenes, and train the top layers on the limited training samples. Ma et al. \cite{ma2016hyperspectral} imposed a relative distance prior on the SAE DL network to deal with training instabilities. This approach extends the SAE by adding the new distance prior term and corresponding SGD optimization. LeCun reviews a number of unsupervised learning algorithms using AEs, which can possibly aid when training data is minimal \cite{lecun2012learning} . Pal \cite{pal2009kernel} reviews kernel methods in RS and argues that SVMs are a good choice when there are a small number of training samples. Petersson et al. \cite{petersson2016hyperspectral} suggest using SAEs to handle small training samples in HSI processing.

%
%
\subsection{Human-understandable solutions for modelling physical phenomena}
\label{subsec:ChallengesOpportunities_ii}
\textbf{Open Question \#2a: How can DL improve model-based RS?} 

Many RS applications depend on models (e.g. a model of crop output given rain, fertilizer and soil nitrogen content, and time of year), many of which are very complicated and often highly nonlinear. Model outputs can be very inaccurate if the models don't adequately capture the true input data and properly handle the intricate inter-relationships between input variables.

Abdel-Rahman et al. \cite{abdel2008application} pointed out that more accurate estimation of nitrogen content and water availability can aid biophysical parameter estimation for improving plant yield models. Ali et al. \cite{Ali201Review} examine biomass estimation, which is a nonlinear and highly complex problem. The retrieval problem is ill-posed and the electromagnetic response is the complex result of many contributions. The data pixels are usually mixed, making this a hard problem. ANNs and support vector regression have shown good results. They anticipate that DL models can provide good results. Both Adam et al. \cite{Adam2010MultiSpectral} and Ozesmi et al. \cite{ozesmi2002satellite} agree that there is need for improvement in wetland vegetation mapping. Wetland species are hard to detect and identify compared to terrestrial plants. Hyperspectral sensing with narrow bandwidths in frequency can aid. Pixel unmixing is important since canopy spectra are similar and combine with underlying hydrologic regime and atmospheric vapor. Vegetation spectra highly correlated among species, making separation difficult. Dorigo et al. \cite{dorigo2007review} analyzed inversion-based models for plant analysis, which is inherently an ill-posed and hard task. They found that using ANN inversion techniques have shown good results. DL may be able to help improve results. Canopy reflections are governed by large number of canopy elements interacting and by external factors. Since DL networks can learn very complex non-linear systems, it seems like there is much room for improvement in applying DL models. DBNs or other DL systems seem like a natural fit for these types of problems.

Kuenzer et al. \cite{kuenzer2014earth} and Wang et al. \cite{wang2010remote} assess biodiversity modeling. Biodiversity occurs at all levels from molecular to individual animals, to ecosystem, to global. This requires a large variety of sensors and analysis at multiple scales. However, a main challenge is low temporal resolution. There needs to be a focus beyond just pixel-level processing, and utilizing spatial patterns and objects. DL systems have been shown to learn hierarchical features, with smaller scale features learned at the beginning of the network, and more complex and abstract features learned in the deeper portions. \\

\textbf{Open Question \#2b: What tools and techniques are required to ``understand'' how the DL works?} 
\ParSection{Human Understandable}

It is also worth mentioning that many of these applications involve biological and scientific end-users, who will definitely want to understand how the DL systems work. For instance, a linear model that models some biological process is easily understood - both the mathematical model and the statistics resulting from estimating the model parameters are well understood by scientists and biologists. However, a DL system can be so large and complex as to defy analysis. We note that this is not specific to RS, but a general problem in the broader DL community.

The DL system is seen by many researchers, especially scientists and RS end-users, as a black box that is hard to understand what is happening ``under the hood". Egmont-Peterson et al. \cite{Egmont-Petersen2002} and Fassnacht et al. \cite{fassnacht2016review} both state that disadvantages of NNs are understanding what they are actually doing, which can be difficult to understand. In many RS applications, just making a decision is not enough; people need to understand how reliable the decision is and how the system arrived at that decision. Ali et al. \cite{Ali2015} also echo this view in their review paper on improving biomass estimation. Visualization tools which show the convolutional filters, learning rates, and tools with deconvolution capabilities to localize the convolutional firings are all helpful \cite{mahendran2015understanding, yosinski2015understanding, zeiler2014visualizing, simonyan2013deep, erhan2009visualizing} . Visualization of what the DL is actually learning is an open area of research. Tools and techniques capable of visualizing what the network is learning and measures of how robust the network is (estimating how well it may genralize) would be of great benefit to the RS community (and the general DL community).

%
%
\subsection{Big Data}
\label{subsec:ChallengesOpportunities_iii}
\textbf{Open Question \#3: What happens when DL meets Big Data?}

As already discussed in Section \ref{subsec:bigdatareview}, a number of mathematics, algorithms and hardware have been put forth to date relative to large scale DL networks and DL in Big Data. However, this challenge is not close to being solved. Most approaches to date have focused on Big Data challenges in RGB or RGBD data for tasks like face and object detection or speech. With respect to remote sensing, we have many of the same problems as CV, but there are unique challenges related to different sensors and data. First, we can break Big Data into its different so-called ``parts'', e.g., volume, variety and velocity. With respect to DBNs, CNNs, AEs, etc., we are primarily concerned with creating new robust and distributed mathematics, algorithms and hardware that can ingest massive streams of large, missing, noisy data from different sources, such as sensors, humans and machines. This means being able to combine image stills, video, audio, text, etc., with symbolic and semantic variation. Furthermore, we require real-time evaluation and possibly online learning. As Big Data in DL is a large topic, we restrict our focus herein to factors that are unique to remote sensing. 

The first factor that we focus on is high spatial and more-so spectral dimensionality. Traditional DLs operate on relatively small grayscale or RGB imagery. However, SAR imagery has challenges due to noise, and MSI and HSI can have from four to hundreds to possibly thousands of channels. As Arel et al. \cite{Arel2010} pointed out, a very difficult question is how well DL architectures scale with dimensionality. To date, preliminary research has tried to combat dimensionality by applying dimensionality reduction or feature selection prior to DL, e.g., Benediktsson et al. \cite{Benediktsson2012} reference different band selection, grouping, feature extraction and subspace identification in HSI remote sensing. 

Ironically, most RS areas suffer from a lack of training data. Whereas they may have massive amounts of temporal and spatial data, there may not be the seasonal variations, times of day, object variation (e.g., plants, crops, etc.), and other factors that ultimately lead to sufficient variety needed to train a DL model. For example, most online hyperspectral data sets have little-to-no variety and it is questionable about what they are, and really can at that, learn. In stark contrast, most DL systems in CV use very large training sets, e.g., millions or billions of faces in different illuminations, poses, inner class variations, etc. Unless the remote sensing DL applies a method like transfer learning, DL in RS often have very limited training data. For example, in \cite{ma2016hyperspectral} Ma et at. tried to address this challenge by developing a new prior to deal with the instability of parameter estimation for HSI classification with small training samples. The SAE is modified by adding the relative distance prior in the fine-tuning process to cluster the samples with the same label and separate the ones with different labels. Instead of minimizing classification error, this network enforces intra-class compactness and attempts to increase inter-class discrepancy. 

%
%
\subsection{Non-traditional heterogeneous data sources}
\label{subsec:ChallengesOpportunities_iv} \textbf{Open Question \#4a: How can DL work with non-traditional data sources?} 

Non-traditional data sources, such a twitter, YouTube, etc. offer data that can be useful to RS. These methods will probably never replace RS, but usually offer benefits to augment RS data, or provide quality real-time data before RS methods, which usually take longer, can provide RS-based data.

Fohringer et al. \cite{Fohringer2015NonTraditional} utilized information extracted from social media photos to enhance RS data for flood assessments. They found one major challenge was filtering posts to a manageable amount of relevant ones to further assess. The data from Twitter and Flickr proved useful for flood depth estimation prior to RS-based methods, which typically take 24-28 hours. Frias-Martinez et al. \cite{Frias-Martinez2014Spectral} take advantage of large amounts of geolocated content in social media by analyzing Twitter tweets as a complimentary source of data for urban land-use planning. Data from Manhattan (New York, USA), London (UK), and Madrid (Spain) was analyzed using a self-organizing map \cite{kohonen1998self} followed by a Voronoi tesselation. Middleton et al. \cite{middleton2014real} match geolocated tweets and created real-time crisis maps via statistical analysis, which are compared to the US National Geospatial Agency post-event impact assessments. A major issue is that only about $1\%$ of tweets contain geolocation data. The tweets usually follow a pattern of a small number of first-hand reports and many re-tweets and comments. High-precision results were obtained. Singh et al. \cite{Singh2010Social} aggregate user's social interest about any particular theme from any particular location into so called ``social pixels'', which are amenable to media processing techniques (e.g., segmentation, convolution), which allow semantic information to be derived. They also developed a declarative operator set to allow queries to visualize, characterize, and analyze social media data. Their approach would be a promising front-end to any social media analysis system. In the survey paper of Sui and Goodchild \cite{Sui2011Convergence} , the convergence of Geographic Information System (GIS) and social media are examined. They observed that GIS has moved from software helping a user at a desk to a means of communicating earth surface data to the masses (e.g. OpenStreetMap, Google Maps, etc.).

In all of the above mentioned methods, DL can play a significant role of parsing data, analyzing data and estimating results from the data. It seems that social media is not going away, and data from social media can often be used to augment RS data in many applications. Thus the question is what novel work awaits the researcher in the area of using DL to combine non-traditional data sources with RS?

\ParSection{Data Fusion}
\textbf{Open Question \#4b: How does DL ingest heterogeneous data?} 

Fusion can take place at numerous so-called ``levels'', including signal, feature, algorithm and decision. For example, Signal In Signal Out (SISO) is where multiple signals are used to produce a signal out. For $\Re$-valued signal data, a common example is the trivial concatenation of their underlying vectorial data, i.e., $X=\{ \hat{x}_1, ... , \hat{x}_N \}$ becomes $[ \hat{x}_1 \hat{x}_2 ... \hat{x}_N ]$ of length $|\hat{x}_1|+...+|\hat{x}_N|$. Feature In Feature Out (FIFO), which is often related to if not the same as SISO, is where multiple features are combined, e.g., a HOG and a Local Binary Pattern (LBP), and the result is a new feature. One example is Multiple Kernel Learning (MKL), e.g., $\ell$-$p$ norm genetic algorithm MKL (GAMKLp) \cite{7762088} . Typically the input is $N$ $\Re$-valued Cartesian spaces and the result is a Cartesian space. Most often, one engages in MKL to search for space in which pattern obey some property, e.g., they are nicely linearly separable and a machine learning tool like a SVM can be employed. On the other hand, Decision In Decision Out (DIDO), e.g., the Choquet integral (ChI), is often used for the fusion of input from \emph{decision makers}, e.g., human experts, algorithms, classifiers, etc. \cite{6722924}. Technically speaking, a CNN is typically a Signal In Decision Out (SIDO) or Feature In Decision Out (FIDO) system. Internally, the \emph{feature learning} part of the CNN is a SIFO or FIFO and the classifier is a FIDO. To date, most DL approaches have ``fused'' via (1) concatenation of $\Re$-valued input data (SISO or FIFO) relative to a single DL, (2) each source has its own DL, minus classification, that is later combined into a single DL, or (3) multiple DLs are used, one for each source, and their results are once again concatenated and subjected to the classifier (either a MLP, SVM or other classifier). 

Herein, we highlight the challenges of syntactic and semantic fusion. Most DL approaches to date syntactically have addressed how $N$ things, which are typically homogeneous mathematically, can be ingested by a DL. However, the more difficult challenge is semantically how should these sources be combined, what is a proper architecture, what is learned (can we understand it) and why should we trust the solution. This is of particular importance to numerous challenges in remote sensing that require a physically meaningful/grounded solution, e.g., model-based approaches. The most typical example of fusion in remote sensing is the combining of data from two (or more) sensors. Whereas there may be semantic variation but little-to-no semantic variation, e.g., both are possibly $\Re$-valued vector data, the reality is most sensors record objective evidence about our universe. However, if human information (e.g., linguistic or text) is involved or algorithmic outputs (e.g., binary decisions, labels/symbols, probabilities, etc.), fusion becomes increasingly more difficult syntactically and semantically. Many theoretical (mathematical and philosophical) investigations, which are beyond the scope of this work, have concerned themselves with how to meaningfully combine objective vs. subjective data/information, qualitative vs. quantitative data/information, or evidences with beliefs other numerous other flavors information. It is a naive and dangerous belief that one can simply just ``cram'' data/information into a DL and get a meaningful and useful result. \textit{How is fusion occurring?} \textit{Where is it occurring?} Fusion is further compounded if one is using uncertain information, e.g., probabilistic, possibilities, or other interval or distribution-based input. The point is, heterogeneous, be it mathematical representation, associated uncertainty, etc., is a real and serious challenge and if the DL community wishes to fuse multiple inputs or sources (humans, sensors and algorithms) then DL must theoretically rise to the occasion to ensure that the architectures and what is subsequently being learned is useful and meaningful. Example, and really preliminary at that, associated DL works to date include fusing hyperspectral with LiDAR \cite{7786851} (two sensors yielding objective data) and text with imagery or video \cite{conficmlNgiamKKNLN11} (thus high-level human information sensor data), to name a few. The point is, the question remains, how can/does DL fuse data/information arising from one or more sources?

%
%
\subsection{DL architectures and learning algorithms for spectral, spatial and temporal data} \label{subsec:ChallengesOpportunities_v} 
\ParSection{Optimal architectures and architectural extensions}
\textbf{DL Open Question \#5: What architectural extensions will DL systems require in order to tackle complicated RS problems?} 

Current DL architectures, components (e.g. convolution), and optimization techniques may not be adequate to solve complex RS problems. In many cases, researchers have developed novel network architectures, new layer structures with their associated SGD or BP equations for training, or new combinations of multiple DL networks. This problem is also an open issue in the broader CV community. This question is at the heart of DL research. Other questions related to the open issues are:

\begin{itemize}
    \setlength{\parskip}{0pt}
    \setlength{\itemsep}{0pt plus 1pt}
    \item \textit{What architecture should be used?} 
    \item \textit{How deep should a DL system be, and what architectural elements will allow it to work at that depth?}
    \item \textit{What architectural extensions (new components) are required to solve this problem?}
    \item \textit{What training methods are required to solve this problem?}
\end{itemize}

We examine several general areas where DL systems have evolved to handle RS data: (i) multi-sensor processing, (ii) utilizing multiple DL systems, (iii) Rotation and displacement-invariant DL systems, (iv) new DL architectures, (v) SAR, (vi) Ocean and atmospheric processing, (vii) 3D processing,  (viii) spectral-spatial processing, and (ix) multi-temporal analysis. Furthermore, we examine some specific RS applications noted in several RS survey papers as areas that DL can benefit: (a) Oil spill detection, (b) pedestrian detection, (c) urban structure detection, (d) pixel unmixing, and (e) road extraction. This is by no means an exhaustive list, but meant to highlight some of the important areas.

\textbf{Multi-Sensor Processing:} Chen et al. \cite{Chen2016DeepFusion} utilize two deep networks, one analyzing HSI pixel neighbors (spatial data), and the other LiDAR data. The outputs are stacked and a FC and logistic regression lay provides outputs. Huang et al. \cite{huang2015new} use a modified sparse DAE (MSDAE) to train the relationship between HR and LR image patches. The stacked MSDAE (S-MSDAE) are used to pretrain a DNN. The HR MSI image is then reconstructed from the observed LR MSI image using the trained DNN.

%
%
\textbf{Multi-DL system:} In certain problems, multiple DL systems can provide significant benefit. Chen et al. \cite{chen2014vehicle} utilize parallel DNNs with no cross-connections to both speed up processing and provide good results in vehicle detection from satellite imagery. Ciresan et al. \cite{Ciresan2012} utilize multiple parallel DNNs that are averaged for image classification. Firth et al. \cite{firth2016novel} use 186 RNNs to perform accurate weather prediction. Hou et al. \cite{Hou2015Polarimetric} use RBMs to train from polarimetric SAR data and a three-layer DBN is used for classification. Kira et al. \cite{Kira2012} used stereo-imaging for robotic human detection, utilizing a CNN which was trained on appearance and stereo disparity-based features, and a second CNN, which is used for long-range detection. Marmanis et al. \cite{Marmanis2016Semantic} utilized an ensemble of CNNs to segment VHR aerial imagery using a FCN to perform pixel-based classification. They trained multiple networks with different initializations and average the ensemble results. The authors also found errors in the dataset, Vaihingen \cite{rottensteiner2012isprs} .

%
%
\textbf{Rotation- and displacement-invariant systems:} Some RS problems require systems that are rotation and displacement-invariant. CNNs have some robustness to translation, but not in general to rotations. Cheng et al. \cite{Cheng2016Learning} incorporated a rotation-invarint layer into a DL CNN architecture to detect objects in satellite imagery. Du et al. \cite{du2016sar} developed a displacement- and rotation-insensitive deep CNN for SAR Automated Target Recognition (ATR) processing that is trained by augmented dataset and specialized training procedure.

%
%
\textbf{Novel DL architectures:} Some problems in RS require novel DL architectures. Dong et al. \cite{dong2016image} use a CNN that takes the LR image and outputs the HR image. He et al. \cite{He2016Hyperspectral} proposed a deep stacking network for HSI classification that utilizes nonlinear activations in the hidden layers and does not require SGD for training. Kontschieder et al. \cite{kontschieder2015deep} developed deep neural decision forests, which uses a stochastic and differentiable decision tree model that steers the representation learning usually conducted in the initial layers of a deep CNN. Lee et al. \cite{Lee2016Contextual} analyze HSI by applying multiple local 3D convolutional filters of different sizes jointly exploiting spatial and spectral features, followed by a fully-convolutional layers to predict pixel classes. Zhang et al. \cite{zhang2016scene} propose GBRCN to classify VHR satellite imagery. Ouyang et al. \cite{ouyang2012discriminative} developed a probabilistic parts-detector based model to robustly handle human detection with occlusions are large deformations utilizing a discriminative RBM to learn the visibility relationship among overlapping parts. The RBM has three layers that handle different size parts. Their results can possibly be improved by adding additional rotation invariance.

%
%
\textbf{Novel DL SAR architectures:} SAR imagery has unique challenges due to noise and the grainy nature of the images. Geng et al. \cite{Geng2015SAR} developed a deep convolutional AE, which is a combination of a CNN, AE, classification, and post-processing layers to classify high-resolution SAR images. Hou et al. \cite{Hou2015Polarimetric} developed a polarimetric SAR DBN. Filters are extracted from the RBMs and a final three-layer DBN performs classification. Liu et al. \cite{Liu2016Terrain} utilize a Deep Sparse Filtering Network to classify terrain using polarimetric SAR data. The proposed network is based on sparse filtering \cite{ngiam2011sparse} , and the proposed network performs a minimization on the output $L_{1}$ norm to enforce sparsity. Qin et al. \cite{qin2017object} performed object-oriented classification of polarimetric SAR data using a RBM and built an adaptive boosting framework (AdaBoost \cite{freund1995desicion} ) vice a stacked DBN in order to handle small training data. They also put forth the RBM-AdaBoost algorithm. Schwegmann et al. \cite{schwegmann2016very} utilized a very deep Highway Network configuration as a ship discrimination stage for SAR ship detection. They also presented a three-class SAR dataset that allows for more meaningful analysis of ship discrimination performances. Zhou et al. \cite{zhao2015three} proposed a three-class change detection approach for multitemporal SAR images using a RBM. These images either have increases or decreases in the backscattering values for changes, so the proposed approach classifies the changed areas into the positive and negative change classes, or no change if none is detected.

%
%
\textbf{Oceanic and atmospheric studies:} Oceanic and atmospheric studies present unique challenges to DL systems that require novel developments. Ducournau et al. \cite{Ducournau2016Deep} developed a CNN architecture, which analyzes sea surface temperature fields and provides a significant gain in terms of peak signal-to-noise ratio compared to classical downscaling techniques. Shi et al. \cite{Shi2015Convolutional} extended the FC-LSTM network that they call ConvLSTM, which has convolutional structures in the input-to-state and state-to-state transitions for precipitation nowcasting.

%
%
\textbf{3D Processing:} Guan et al. \cite{Guan2015Deep} use voxel-based filtering removes ground points from LiDAR data, then a DL architecture generates high-level features from the trees’ 3D geometric structure. Haque et al. \cite{Haque2016} utilize both of CNN and RNN to process 4D spatio-temporal signatures to idenify humans in the dark.

%
%
\textbf{Spectral-spatial HSI processing:} HSI processing can be improved by fusion of spectral and spatial information. Ma et al. \cite{ma2016spectral} propose a spatial updated deep AE which adds a similarity regularization term to the energy function to enforce spectral similarity. The regularization terms is a a cosine similarity term (basically the spectral angle mapper) between the edges of a graph, which the nodes are samples, which enforces keeping the sample correlations. Ran et al. \cite{ran2016bands} classify HSI data by learning multiple  CNN-based submodels for each correlated set of bands, while in parallel a conventional CNN learns spatial-spectral characteristics. The models are combined at the end. Li et al. \cite{li2017hyperspectral} incorporated vicinal pixel information by combining the center pixel and vicinal pixels, and utilizing a voting strategy to classify the pixels. 

%
%
\textbf{Multi-temporal analysis:} Multi-temporal analysis is a subset of RS analysis that has its own challenges. Data revisit rates are often long, and ground-truth data is even more expensive as multiple imagery sets have to be analyzed, and images must be co-registered for most applications.

Jianya et al. \cite{jianya2008review} review multi-temporal analysis, and observe that it is hard, the changes are often non-linear, and changes occur on different timescales (seasons, weeks, years, etc.). The process from ground objects to images is not reversible, and image change to earth change is a very difficult task. Hybrid method involving classification, object analysis, physical modeling, and time series analysis can all potentially benefit from DL approaches. Arel et al. \cite{Arel2010} ask if DL frameworks can understand trends over short, medium and long times? This is an open question for RNNs.

Change detection is an important subset of multi-temporal analysis. Hussain et al. \cite{hussain2013change} state that change detection can benefit from texture analysis, accurate classifications, and ability to detect anomalies. DL has huge potential to address these issues, but it is recognized that DL algorithms are not common in image processing software in this field and large training sets and large training times may also be required. In cases of non-normal distributions, ANNs have shown superior results to other statistical methods. They also recognize that DL-based change detection can go beyond traditional pixel-based change detection methods. Tewkesbury et al. \cite{tewkesbury2015critical} observe that change detection can occur at the pixel, kernel (group of pixels), image-object, multi-temporal image-object (created by segmenting over time series), vector-polygon, and hybrid. While the pixel level is suitable for many applications, hybrid approaches can yield better results in many cases. Approaches to change detection can utilize DL to (1) co-register images and (2) detect changes at hierarchical (e.g. more than just pixel levels).

%
%
\textbf{Some selected specific applications that can benefit from DL analysis:} This section discusses some selected applications where DL can benefit the results. This is by no means an exhaustive list, and many other areas can potentially benefit from DL approaches. In oil spill detection, Brekke et al. \cite{brekke2005oil} point out that training data is scarce. Oil spills are very rare, which usually means oil spill detection approaches are anomaly detectors. Physical proximity, slick shape, and texture play important roles. SAR imagery is very useful, but there are look-alike phenomena that cause false positives. Algal information fusion from optical sensors and probability models can aid detection. Current algorithms are not reliable, and DL has great promise in this area.

In the area of pedestrian detection, Dollar et al. \cite{dollar2012pedestrian} discuss that many images with pedestrians have only a small number of pixels. Robust detectors must handle occlusions. Motion features can achieve very high performance, but few have utilized them. Context (ground plane) approaches are needed, especially at lower resolutions. More datasets are needed, especially with occlusions. Again, DL can provide significant results in this area.

For urban structure analysis, Mayer et al. \cite{mayer1999automatic} report that scale-space analysis is required due to different scales of urban structures. Local contexts can be utilized in the analysis. Analyzing parts (dormers, windows, etc) can improve results. Sensor fusion can aid results. Object variability is not treated sufficiently (e.g. highly non-planar roofs). The DL system's ability to learn hierarchical components and learn parts makes is a good candidate for improving results in this area.

In pixel unmixing, Shi et al. \cite{shi2014incorporating} and Somers et al. \cite{somers2011endmember} review papers both point out that whether an unmixing system uses a spectral library or extracts endmembers spectra from the imagery, the accuracy highly depends on the selection of appropriate endmembers. Adding information from a spatial neighborhood can enhance the unmixing results. DL methods such as CNNs or other tailored systems can potentially inherently combine spectral and spatial information. DL systems utilizing denoising, iterative unmixing, feature selection, spectral weighting, and spectral transformations can benefit unmixing.

Finally in the area of road extraction, Wang et al. \cite{wang2016review} point out that roads can have large variability, are often curves, and can change size. In bad weather, roads can be very hard to identify. Object shadows, occlusions, etc. can cause the road segmentation to miss sections. Multiple models and multiple features can improve results. The natural ability of DL to learn complicated hierarchical features from data makes them a good candidate for this application area also.

%
%
\subsection{Transfer Learning} \label{subsec:ChallengesOpportunities_vi} \textbf{Open Question \#6: How can DL in RS successfully utilize transfer learning?}

In general, we note that transfer learning is also an open question in DL in general, not just in DL related to remote sensing. Section \ref{subsec:ChallengesOpportunities_ix} discusses transfer learning in the broader contect of the entire field of DL, which this section discusses transfer learning in a RS context.

According to Tuia et al. \cite{tuia2016domain} and Pan et al. \cite{pan2010survey} , transfer learning seeks to learn from one area to another in one of four ways: instance-transfer, feature-representation transfer, parameter transfer, and relational-knowledge transfer. Typically in remote sensing, when changing sensors or changing to a different part of a large image or other imagery collected at different times, the transfer fails. Remote sensing systems need to be robust, but doesn't necessarily require near-perfect knowledge. Transfer between HSI images where the number and types of endmembers are different has very few studies. Ghazi et al. \cite{ghazi2017plant} suggest that two options for transfer learning are to (1) utilize pre-trained network and learn new features in the imagery to be analyzed, or (2) fine-tune the weights of the pre-trained network using the imagery to be analyzed. The choice depends on the size and similarity of the training and testing datasets.

There are many open questions about transfer learning in HSI RS:
\begin{itemize}
    \setlength{\parskip}{0pt}
    \setlength{\itemsep}{0pt plus 1pt}
    \item \textit{How does HSI transfer work when the number and type of endmembers are different?}
    \item \textit{How can DL systems transfer low-level to mid-level features from other domains into RS?}
    \item \textit{How can DL transfer learning be made robust to imagery collected at different times and under different atmospheric conditions?}
\end{itemize}

Although in general these open questions remain, we do note that the following papers have successfully utilized transfer learning in RS applications: Yang et al. \cite{Yang2016Two} trained on other remote sensing imagery and transferred low-level to mid-level features to other imagery. Othman et al. \cite{Othman2016Using} utilized transfer learning by training on the ILSVRC-12 challenge data set, which has 1.2 million $224 \times 224$ RGB images belonging to 1,000 classes. The trained system was applied to the UC Merced Land Use \cite{yang2010bag} and Banja-Luka \cite{risojevic2011gabor} datasets. Iftene et al. \cite{iftene2016very} applied a pretrained CaffeNet and GoogleNet models on the ImageNet dataset, and then applying the results to the VHR imagery denoted the WHU-RS dataset \cite{chatfield2014return, sheng2012high}. Xie et al. \cite{xie2015transfer} trained a CNN on night-time imagery and used it in a poverty mapping. Ghazi et al. \cite{ghazi2017plant} and Lee et al. \cite{Lee20171} used a pre-trained networks AlexNet, GoogLeNet and VGGNet on the LifeCLEF 2015 plant task dataset \cite{joly2016lifeclef} and MalayaKew dataset \cite{LeeCWR15} for plant identification. Alexandre \cite{Alexandre2016} used four independent CNNs, one for each channel of RGBD, instead of using a single CNN receiving the four input channels. The four independent CNNs are then trained in a sequence by using the weights of a trained CNN as starting point to train the other CNNs that will process the remaining channels. Ding et al. \cite{ding2016deep} utilized transfer learning for automatic target recognition from mid-wave infrared (MWIR) to longwave IR (LWIR). Li et al. \cite{Li2017Anomaly} used transfer learning by utilizing pixel-pairs based o reference data with labeled sampled using Airborne Visible / Infrared Imaging Spectrometer (AVIRIS) hyperspectral data.

%
%
\subsection{An improved theoretical understanding of DL systems}
\label{subsec:ChallengesOpportunities_vii} 
\textbf{DL Open Question \#7: What new developments will allow researchers to better understand DL systems theoretically?}

The CV and NN image processing communities understand BP and SGD, but until recently, researchers struggled to train deep networks. One issue has been identified as vanishing or exploding gradients \cite{bengio1994learning, glorot2010understanding} . Using normalized initialization and normalization layers can help alleviate this problem. Using special architectures, such as deep residual learning \cite{he2016deep} or highway networks \cite{srivastava2015training} feed data into the deeper layers, thus allowing very deep networks to be trained. FC networks \cite{long2015fully} have achieved success in pixel-based semantic segmentation tasks, and are another alternative to going deep. Sokoli\'{c} et al. \cite{sokolic2016robust} determined that the spectral norm of the NN's Jacobian matrix in the neighbourhood of the training samples must be bounded in order for the network to generalize well. All of these methods deal with a central problem in training very deep NNs: The gradients must not vanish, explode, or become too uncorrelated, or else learning is severely hindered.

The DL field needs practical (and theoretical) methods to go deep, and ways to train efficiently with good generalization capabilities. Many DL RS systems will probably require new components, and these networks with the new components need to be analyzed to see if the methods above (or new methods not yet invented) will enable efficient and robust network training. Egmont-Peterson et al. \cite{Egmont-Petersen2002} point out that DL training is sensitive to the initial training samples, and it is a well-known problem in SGD and BP of potentially reaching a local minimum solution but not being at the global minimum. In the past, seminal papers such as Hinton's \cite{hinton2006fast} which allow efficient training of the network, allow researchers to break past a previously difficult barrier. What new algorithmic and theoretical developments will spur the next large surge in DL?

%
%
\subsection{High barriers to entry}
\label{subsec:ChallengesOpportunities_viii} 
\textbf{DL Open Question \#8: How to best handle high entry barriers to DL?} 

Most DL papers assume that the reader is familiar with DL concepts, backpropagation, etc. This is in reality a steep learning curve that takes a long time to master. Good tutorials and online training can aid students and practitioners who are willing to learn. Implementing BP or SGD on a large DL system is a difficult task, and simple errors can be hard to determine. Furthermore, BP can fail in large networks, so alternate architectures such as highway nets are usually required.

Many DL systems have a large number of parameters to learn, and often require large amounts of training data. Computers with GPUs and GPU-capable DL programs can greatly benefit by offloading computations onto the GPUs. However, multi-GPU systems are expensive, and students often use laptops that cannot be equipped with a GPU. Some DL systems run under Microsoft Windows, while others run under variants of Linux (e.g. Ubuntu or Red Hat). Futhermore, DL systems are programmed in a variety of languages, including Matlab, C, C++, Lua, Python, etc. Thus practitioners and researchers have a potentially steep learning curve to create custom DL solutions.

Finally, the large variety of data types in remote sensing, including RGB imagery, RGBD imagery, MSI, HSI, SAR, LiDAR, stereo imagery, tweets, GPS data, etc., all of which may require different architectures of DL systems. Often, many of the tasks in the RS community require components that are not part of a standard DL library tool. A good understanding of DL systems and programming is required to integrate these components into off-the-shelf DL systems.

%
%

\subsection{Training}
\label{subsec:ChallengesOpportunities_ix} 
\textbf{Open Question \#9: How to train and optimize the DL system?} 

Training a DL system can be difficult. Large systems can have millions of parameters. There are many methods that DL researchers use to effectively train systems. These methods are discussed below.

\textbf{Data imputation:} Data imputation \cite{hinton2006fast} is important in RS, since there are often a small number of training samples. In imagery, image patched can be extracted and stretched with affine transformations, rotated, and made lighter or darker (scaling). Also, patched can be zeroed (removed) from training data to help the DL be more robust to occlusions. Data can also be augmented by simulations. Another method that can be useful in some circumstances is domain transfer, discussed below (transfer learning).
    
\textbf{Pre-training:} Erhan et al. \cite{Erhan2010} performed a detailed study trying to answer the questions ``\textit{How does unsupervised pre-training work?}'' and ``\textit{Why does unsupervised pre-training help DL?}''. Their empirical analysis shows that unsupervised pre-training guides the learning towards attraction basins of minima that support better generalization and pre-training also acts as a regularizer. Furthermore, early training example have a large impact on the overall DL performance. Of course, these are experimental results, and results on other datasets or using other DL methods can yield different results. Many DL systems utilize pre-training followed by fine-tuning.
    
\textbf{Transfer Learning:} Transfer learning is also discussed in section \ref{subsec:ChallengesOpportunities_vi}. Transfer learning attempts to transfer learned features (which can also be thought of as DL layer activations or outputs) from one image to another, from one part of an image to another part, or from one sensor to another. This is a particularly thorny issue in RS, due to variations in atmosphere, lighting conditions, etc. Pan et al. \cite{pan2010survey} point out that typically in remote sensing, when changing sensors or changing to a different part of a large image or other imagery collected at different times, the transfer fails. Remote sensing systems need to be robust, but they don't necessarily require near-perfect knowledge. Also, transfer between images where the number and types of endmembers are different has very few studies. Zhang et al. \cite{Zhang2016} also cite transfer learning as an open issue in DL in general, and not just in RS.
    
\textbf{Regularization:} Regularization is defined by Goodfellow et al. \cite{goodfellow2016deep} as ``any modification we make to a learning algorithm that is intended to reduce its generalization error but not its training error.'' There are many forms of regularizer - parameter size penalty terms (such as the $L_{2}$ or $L_{1}$ norm, and other regularizers that enforce sparse solutions; diagonal loading of a matrix so the matrix inverse (which is required for some algorithms) is better conditioned; Dropout and early stopping (both are described below); adding noise to weights or inputs; semi-supervised learning, which usually means that some function that has a very similar representation to examples from the same class is learned by the NN; Bagging (combining multiple models); and adversarial training, where a weighted sum of the sample and an adversarial sample is used to boost performance. The interested reader is referred to chapter 7 of \cite{goodfellow2016deep} for further information. An interesting DL example in RS is, Mei et al. \cite{Mei2016Integrating} , who utilized a Parametric Rectified Linear unit (PReLu) \cite{he2015delving} , which can help improve model fitting without adding computational cost and with little overfitting risk.
        
\textbf{Early stopping:} Early stopping is a method where the training validation error is monitored and previous coefficient values are recorded. Once the training level reaches a stopping criteria, then the coefficients are used. Early stopping helps to mitigate overtraining. It also acts as a regularizer, constraining the parameter space to be close to the initial configuration \cite{Erhan2010} .
        
\textbf{Dropout:} Dropout usually uses some number of randomly selected links (or a probability that a link will be dropped) \cite{Srivastava2014Dropout} . As the network is trained, these links are zeroed, basically stopping data from flowing from the shallower to deeper layers in the DL system. Dropout basically allows a bagging-like effect, but instead of the individual networks being independent, they share values, but the mixing occurs at the dropout layer, and the individual subnetworks share parameters \cite{goodfellow2016deep} .
    
\textbf{Batch Normalization:} Batch normalization was developed by Ioffe et al. \cite{ioffe2015batch}. Batch normalization breaks the data into small batches and then normalizes the data to be zero mean and unity variance. Batch normalization can also be added internally as layers in the network. Batch normalization reduces the so-called \textit{internal covariate shift} problem for each training mini-batch. Applying batch normalization had the following benefits: (1) allowed a higher learning rate; (2) the DL network was not as sensitive to initialization; (3) dropout was not required to mitigate overfitting; (4) The $L_{2}$ weight regularization could be reduced, increasing accuracy. Adding batch normalization increases the two extra parameters per activation. 
           
\textbf{Optimization:} Optimization of DL networks is a major area of study in DL. It is nontrivial to train a DL netowrk, much less squeeze out high performance on both the training and testing datasets. SGD is a training method that uses small batches of training data to generate an estimate of the gradients. Li et al. \cite{Le2011Optimization} argues the SGD is not inherently parallel, and often requires training many models and choosing the one that performs best on the validation set. They also show that no one method works best in all cases. They found that optimization performance varies per problem. A nice review paper for many gradient descent algorithms is provided by Ruder \cite{ruder2016overview} . According to Ruder, complications for gradient descent algorithms include: 

\begin{itemize}
    \setlength{\parskip}{0pt}
    \setlength{\itemsep}{0pt plus 1pt}
    \item \textit{How to choose a proper learning rate?} 
    \item \textit{How to properly adjust learning-rate schedules for optimal performance? }
    \item \textit{How to adjust learning rates independently for each parameter?}
    \item \textit{How to avoid getting trapped in local minima and saddle points when one dimension slopes up and one down (the gradients can get very small and training halts)}
\end{itemize}

Various gradient descent methods such as Adagrad \cite{duchi2011adaptive} , which adapts the learning rate to the parameters, AdaDelta \cite{zeiler2012adadelta} , which uses a fixed size window of past data, and Adam \cite{kingma2014adam} , which also has both mean and variance terms for the gradient descent, can be utilized for training. Another recent approach seeks to optimize the learning rate from the data is described in Schaul et al. \cite{Schaul2013NoMore} . Finally, Sokoli\'{c} et al. \cite{sokolic2016robust} concluded experimentally that for a DNN to generalize well, the spectral norm of the NN's Jacobian matrix in the neighbourhood of the training samples must be bounded. They furthermore show that the generalization error can be bounded independent of the DL network's depth or width, provided that the Jacobian spectral norm is bounded. They also analyze residual networks, weight normalized networks, CNN's with batch normalization and Jacobian regularization, and residual networks with Jacobian regularization. The interested reader is referred to chapter 8 of \cite{goodfellow2016deep} for further information.

\textbf{Data Propagation:} Both highway networks \cite{srivastava2015highway} and residual networks \cite{he2015deep} are methods that take data from one layer and incorporate it, either directly (highway networks) or as a difference (residual networks) into deeper layers. These methods both allow very deep networks to be trained, at the expense of some additional components. Balduzzi et al. \cite{balduzzi2016neural} examined networks and determined that there is a so-called ``shattered gradient'' problem in DNN, which is manifested by the gradient correlation decaying exponentially with depth and thus gradients resemble white noise. A ``looks linear'' initialization is developed that prevents the gradient shattering. This method appears not to require skip connections (highway networks, residual networks).

%
%
\section{Conclusions}
\label{sec:Conclusions}

In this letter, we have performed a thorough review and analyzed 207 RS papers that utilize FL and DL, as well as 57 survey papers in DL and RS. We provide researches with a clustered set of 12 areas where DL RS papers have been applied. We examine why DL is popular and what is enabling DL. We examined many DL tools and provided opinions about the tools pros and cons. We critically looked at the DL RS field and identified nine general areas with unsolved challenges and opportunities, specifically enumerated 11 difficult and thought-provoking open questions in this area. We reviewed current DL research in CV and discussed recent methods that could be utilized in DL in RS. We provide a table of DL survey papers covering DL in RS and feature learning in RS.

\subsection*{Disclosures}
The authors declare no conflict of interest.

%
%
\acknowledgments 
The authors wish to thank graduate students Vivi Wei, Julie White and Charlie Veal for their valuable inputs related to DL tools. 

%
%
\bibliography{refs}   
\bibliographystyle{spiejour}   

%
%
\vspace{2ex}\noindent\textbf{John E. Ball} is an Assistant professor of Electrical and Computer Engineering at Mississippi State University (MSU), USA. He received the Ph.D. degree in in Electrical Engineering from Mississippi State University in 2007, with a certificate in remote sensing. He is a co-director of the Sensor Analysis and Intelligence Laboratory (SAIL) at MSU and the director of the Simrall Radar Laboratory. He is the author of 45 journal and conference papers, and 22 technical tutorials, white papers, and technical reports, and has written one book chapter. He received best research paper of the year from the Veterinary and Comparative Orthopaedics and Traumatology in 2016 and technical paper of the year award from the Georgia Tech Research Institute in 2012. His current research interests include deep learning, remote sensing, remote sensing, machine learning, digital signal and image processing, and radar systems. Dr. Ball is an associate editor for the SPIE Journal of Applied Remote Sensing. 

\vspace{1ex}
\vspace{2ex}\noindent\textbf{Derek T. Anderson} received the Ph.D. in electrical and computer engineering (ECE) in 2010 from the University of Missouri, Columbia, MO, USA. He is currently an Associate Professor and the Robert D. Guyton Chair in ECE at Mississippi State University (MSU), USA, an Intermittent Faculty Member with the Naval Research Laboratory, co-director of the Sensor Analysis and Intelligence Laboratory (SAIL) at MSU and an Associate Editor for the IEEE Transactions on Fuzzy Systems. His research interests include new frontiers in data/information fusion for pattern recognition and automated decision making in signal/image understanding and computer vision with an emphasis on uncertainty and heterogeneity. Prof. Anderson's primary research contributions to date include multi-source (sensor, algorithm and human) fusion, Choquet integrals (extensions, embeddings, learning), signal/image feature learning, multi-kernel learning, cluster validation, hyperspectral image understanding and linguistic summarization of video. He has published 100+ (journal, conference and book chapter) articles, he is the program co-chair of FUZZ-IEEE 2019, he co-authored the 2013 best student paper in Automatic Target Recognition at SPIE, he received the best overall paper award at the IEEE International Conference on Fuzzy Systems (FUZZ-IEEE) 2012, and he received the 2008 FUZZ-IEEE best student paper award.

\vspace{1ex}
\vspace{2ex}\noindent\textbf{Chee Seng Chan} received the Ph.D. degree from the University of Portsmouth, Hampshire, U.K., in 2008. He is currently a Senior Lecturer with the Department of Artificial Intelligence, Faculty of Computer Science and Information Technology, University of Malaya, Kuala Lumpur, Malaysia. His current research interests include computer vision and fuzzy qualitative reasoning, with an emphasis on image and video understanding. Dr. Chan was a recipient of the Institution of Engineering and Technology (Malaysia) Young Engineer Award in 2010, the Hitachi Research Fellowship in 2013, and the Young Scientist Network-Academy of Sciences Malaysia in 2015. He is the Founding Chair of the IEEE Computational Intelligence Society, Malaysia Chapter, and the Founder of Malaysian Image Analysis and Machine Intelligence Association. He is a Chartered Engineer of the Institution of Engineering and Technology, U.K.

\listoffigures
\listoftables

\end{spacing}
\end{document}